\documentclass[journal]{IEEEtran}

\usepackage[table]{xcolor}
\usepackage{colortbl}
\usepackage[pdftex]{graphicx}
\graphicspath{{../pdf/}{../jpeg/}}
\DeclareGraphicsExtensions{.pdf,.jpeg,.png}

\usepackage[cmex10]{amsmath}

\usepackage{booktabs}       
\usepackage{makecell}       
\usepackage{microtype}      
\usepackage{array}
\usepackage{mdwmath}
\usepackage{hyperref}       
\usepackage{amsmath,amsthm,amssymb,amsfonts}
\usepackage{mdwtab}

\usepackage{eqparbox}
\usepackage{lipsum}
\usepackage{url}
\usepackage{ulem}
\usepackage{algorithm}
\usepackage{algorithmic}

\usepackage{bbding}
\usepackage{multirow}

\begin{document}
\bstctlcite{IEEEexample:BSTcontrol}
    \title{SGCCNet: Single-Stage 3D Object Detector With Saliency-Guided Data Augmentation and Confidence Correction Mechanism}
  \author{Ao Liang, Wenyu Chen, Jian Fang, Huaici Zhao*
  \thanks{Manuscript received April 9, 2024. This work was funded in part by CAS Innovation Fund, under Award Number E01Z040101.}
\thanks{Ao Liang, Wenyu Chen and Huaici Zhao are with the Key Laboratory of Opto-Electronic Information Processing, Chinese Academy of Sciences, Shenyang 110016, China. Shenyang Institute of Automation, Chinese Academy of Sciences, Shenyang 110016, China. University of Chinese Academy of Sciences, Beijing 100049, China, and also with Key Laboratory of Optical Information and Simulation Technology, Liaoning Province, Shenyang 110016, China(e-mail: \{liangao,chenwenyu,hczhao\}@sia.cn).}
\thanks{Jian Fang is with the Key Laboratory of Opto-Electronic Information Processing, Chinese Academy of Sciences, Shenyang 110016, China. Shenyang Institute of Automation, Chinese Academy of Sciences, Shenyang 110016, China, and also with Key Laboratory of Optical Information and Simulation Technology, Liaoning Province, Shenyang 110016, China(e-mail: jianfang@sia.cn).}%
\thanks{Huaici Zhao is the corresponding author.}}

\markboth{IEEE Transactions on Intelligent Vehicles, VOL.~xx, NO.~xx, DECEMBER~2024
}{Liang \MakeLowercase{\textit{et al.}}: Single-Stage 3D Object Detector}

\maketitle

\begin{abstract}
The single-stage point-based 3D object detectors have attracted widespread research interest due to their advantages of lightweight and fast inference speed. However, they still face challenges such as inadequate learning of low-quality objects (ILQ) and misalignment between localization accuracy and classification confidence (MLC). 
In this paper, we propose SGCCNet to alleviate these two issues. For ILQ, SGCCNet adopts a Saliency-Guided Data Augmentation (SGDA) strategy to enhance the robustness of the model on low-quality objects by reducing its reliance on salient features. Specifically, We construct a classification task and then approximate the saliency scores of points by moving points towards the point cloud centroid in a differentiable process. During the training process, SGCCNet will be forced to learn from low saliency features through dropping points. 
Meanwhile, to avoid internal covariate shift and contextual features forgetting caused by dropping points, we add a geometric normalization module and skip connection block in each stage. 
For MLC, we design a Confidence Correction Mechanism (CCM) specifically for point-based multi-class detectors. This mechanism corrects the confidence of the current proposal by utilizing the predictions of other key points within the local region in the post-processing stage. 
Extensive experiments on the KITTI dataset demonstrate the generality and effectiveness of our SGCCNet. On the KITTI \textit{test} set, SGCCNet achieves $80.82\%$ for the metric of $AP_{3D}$ on the \textit{Moderate} level, outperforming all other point-based detectors, surpassing IA-SSD and Fast Point R-CNN by $2.35\%$ and $3.42\%$, respectively. Additionally, SGCCNet demonstrates excellent portability for other point-based detectors.
\end{abstract}


%
\IEEEpeerreviewmaketitle


\section{Introduction}
\IEEEPARstart{W}{ith} the rapid development of intelligent transportation and robot technology, LiDAR-based 3D object detection has become a key technology for intelligent agents to acquire environmental information, attracting widespread research interest \cite{wang2023multi, meng2023hydro}. Within related methods, single-stage point-based detectors can achieve a better balance between accuracy and inference efficiency, gaining increasing attention \cite{yang20203dssd, chen2022sasa, huang2023object, zhang2022not}.

\begin{figure}[t]
	\begin{center}
	\includegraphics[width=3in]{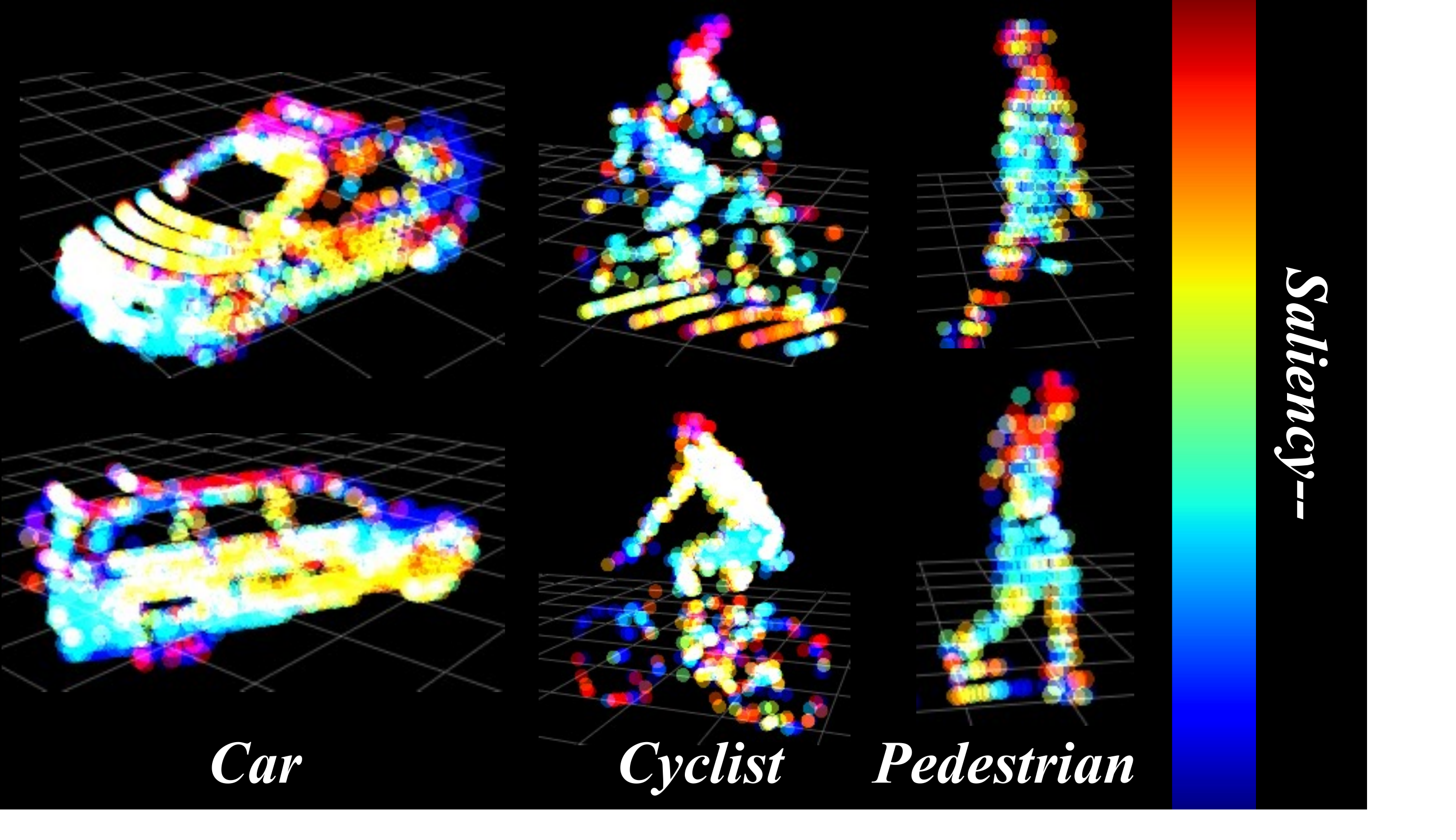}
	\end{center}
	\vspace{-0.3cm}
	\caption{Visualize the saliency of three classes of objects in KITTI. The model's reliance on highly salient features is detrimental to the detection of low-quality objects.}
	\label{fig1}
	\vspace{-0.3cm}
\end{figure}

For example, IA-SSD \cite{yang2022dbq} and DBQ-SSD \cite{yang2022dbq} can respectively achieve impressive 82FPS and 162FPS inference efficiency on the KITTI dataset \cite{Geiger2012CVPR} while maintaining ideal detection performance, surpassing other structured point cloud detectors comprehensively, greatly enhancing the application prospects of point-based detectors on devices with high real-time requirements. However, these type detectors still suffer from two prominent issues, namely Insufficient Learning of Low-Quality Objects (ILQ) and Misalignment between Localization Accuracy and Classification Confidence (MLC).

For ILQ, as limited training data cannot cover all possible feature distributions, especially for low-quality targets that appear less frequently, detectors will lack learning about them. And under the paradigm of repeated sampling and limited augmentation \cite{s18103337}, the models will develop feature dependencies, giving better decision-making capabilities to prominent features while neglecting the learning of other features. In SPSNet, Liang et al. \cite{liang2023spsnet} obtained a way to represent point saliency through loss competition. When only discarding the top 10 most salient points, the model's average $AP_{3D}$ decreased by 8.61\%, providing an intuitive explanation for this scenario. As shown in Fig. \ref{fig1}, we visualize the saliency scores of several objects in KITTI dataset. It can be seen that the distribution of saliency features is relatively fixed, with \textit{Car} objects concentrated near the roof, \textit{Cyclist} objects near the head and wheels, and \textit{Pedestrian} objects near the head and feet. For low-quality objects, these strong saliency distribution areas are likely to be missing and sparse. If the model's feature learning capability is limited to saliency regions, it will greatly reduce its robustness.

\begin{figure}[t]
	\begin{center}
		\includegraphics[width=3.5in]{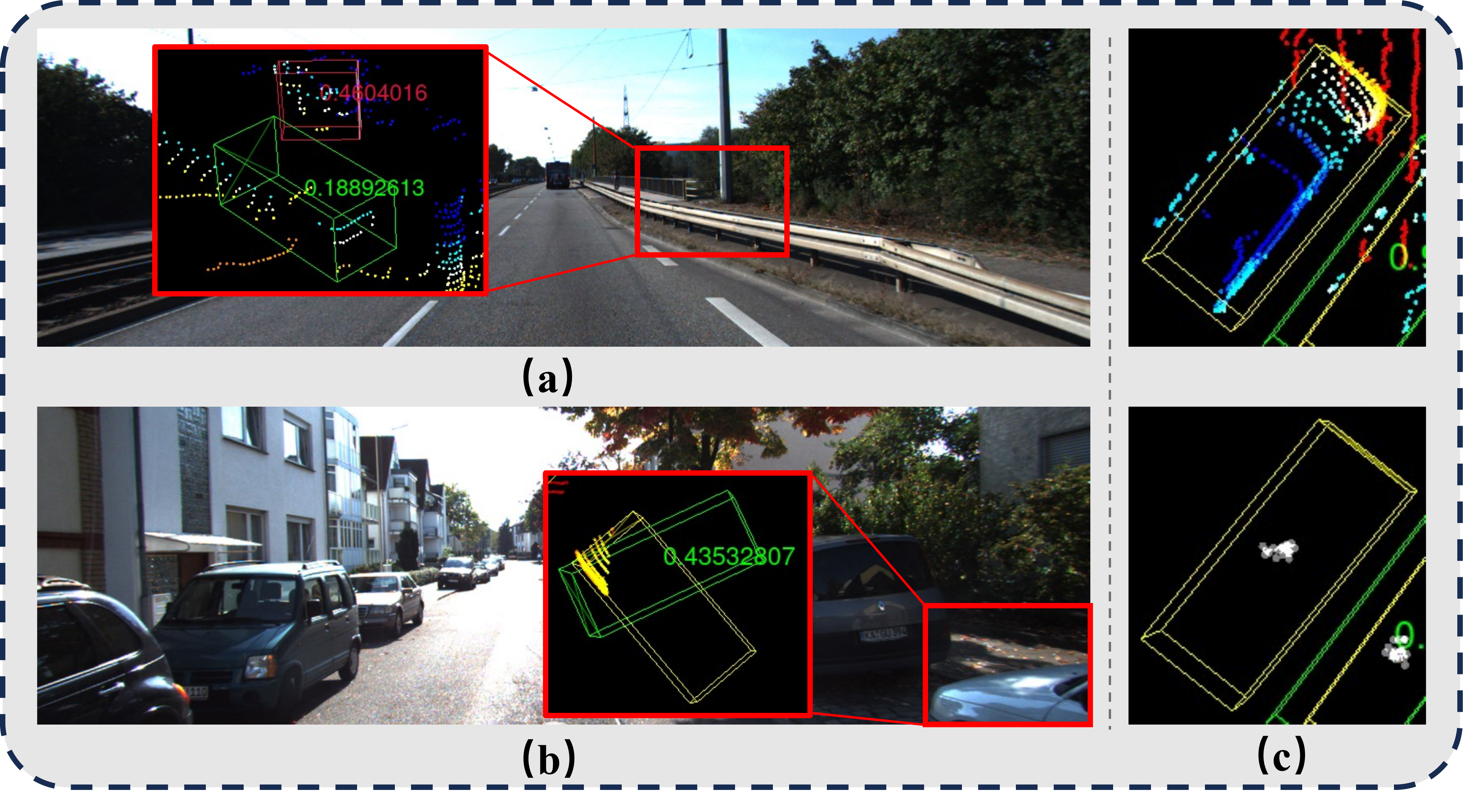}
	\end{center}
	\vspace{-0.3cm}
	\caption{Three typical scenarios of MLC in point-based single-stage 3D detectors. (a) False positive targets. (b) Suboptimal predicted boxes. (c) Missed accurately located targets.}
	\label{fig2}
	\vspace{-0.3cm}
\end{figure}

The MLC problem is inherent in single-stage detectors, unlike two-stage detectors which can refine Rigions of Interesting (RoIs) for a second time. The decision on target confidence and geometric information regression is determined by two separate network branches without any connection between them. This can mislead the Non Max Suppression (NMS) operation in post-processing. We list three typical examples in Fig. \ref{fig2}. The first is the detection of too many false positive targets, where the detection results overly rely on the model's semantic learning ability based solely on confidence. This is disadvantageous for detectors using point-based 3D backbones because point clouds are geometrically rich but textureless, leading to a high probability of objects with similar geometry to be detected as false positives. The second scenario is when the NMS selects target box positions and sizes that are not optimal. Although the selected points have the highest confidence, their IoU with GT is lower than other key points. In the third scenario, the accurate target position can be accurately predicted by the model, but it is not classified as a foreground category. All three scenarios directly reduce the model's performance metrics.

In this paper, we propose SGCCNet to alleviate the impact of the above two issues. Firstly, SGCCNet employs a saliency-guide data augmentation (SGDA) method to enrich the feature distribution of foreground objects in the training data. Specifically, We use a differentiable process of moving points towards the point cloud centroid to act as a point dropout process to approximate the saliency scores of points. During the learning process, we randomly drop a certain number of salient points of each object to create a new ground truth for GT sampling.
For the MLC problem, we have designed a Confidence Correction Mechanism (CCM) in the post-processing stage. For proposals with low IoU, CCM reduces their confidence. For proposals that are considered background by the model but have a significant overlap with bounding boxes in the neighborhood, CCM increases their probability of being true positive ones.

We validated the superiority of SGCCNet on the KITTI dataset. Specifically, the proposed SGCCNet outperforms all previously published point-based approaches. Under the multi-class training scheme, SGCCNet achieved 80.82\% $AP_{3D}$ on the \textit{Car} target in the KITTI \textit{test} set, surpassing prior methods that utilize structure-based backbones while also demonstrating higher efficiency. For the \textit{Pedestrian} and \textit{Cyclist} targets, compared to the latest point-based models, SGCCNet also showed improvements of 1\% and 3.1\% $AP_{3D}$ respectively. In summary, the key contributions of our work are as follows:

\begin{itemize}
	\item A new saliency-guided data augmentation method is proposed, which enhances the diversity of the training data distribution at the feature level, thus improving the robustness of the model to low-quality object detection.
	\item A new point-based backbone that combines geometric normalization modules and skip connection blocks is proposed to alleviate the internal covariate shift problem and feature forgetting problem.
	\item A confidence correction mechanism for post-processing is proposed to effectively address the misalignment between localization accuracy and classification confidence (MLC) issue commonly seen in single-stage detectors.
	\item A high-efficiency and high-precision single-stage point-based 3D detector SGCCNet is proposed, and experimental results show that our method outperforms other similar methods significantly on the KITTI dataset.
\end{itemize}

The structure following this text is as follows: In Section \ref{sec:2}, we review the design principles of classic and state-of-the-art 3D object detectors, and list some methods to overcome ILQ and MLC problems. In Section \ref{sec:3}, we introduce the key components of SGCCNet, detail the implementation process of saliency-guide data augmentation, and finally introduce its end-to-end training loss. In Section \ref{sec:4}, we demonstrate the superiority of SGCCNet through comprehensive comparative experiments and ablation studies. Finally, in Section \ref{sec:5}, we summarize this paper and provide prospects for future work.

\section{Related Work}
\label{sec:2}
As mentioned above, LiDAR-based 3D detectors can be classified into grid-based, range-based, and point-based according to the data form before inputting to models. In this section, we will review the key ideas of classic and state-of-the-art models within each domain, and analyze in detail the measures currently taken by models to alleviate the issues of ILQ and MLC.

\subsection{3D Detectors}
\textbf{Grid-based.}
There are three major types of grid representations: voxels, pillars, and BEV feature maps. Voxel-based models \cite{zhou2018voxelnet,chen2023voxelnext} can preserve the spatial information of the original point cloud scene to the greatest extent based on the grid size. Pillar-based methods \cite{lang2019pointpillars,li2023pillarnext} normalize the voxel space along the z-axis and significantly reducing the number of cells.
The Bird’s-eye view (BEV) feature map is a dense 2D representation, where each pixel corresponds to a specific region and encodes the points information in this region \cite{yin2021center,zhang2024hednet, shi2020pv, shi2023pv, paigwar2021frustum}. In recent years, researchers have found that BEV representation can naturally integrate with downstream tasks such as behavior decision-making and trajectory planning to form end-to-end intelligent systems, leading to a resurgence in research on perception models based on BEV \cite{10321736, li2022bevformer, li2023lanesegnet, zeng2023distilling}.

\textbf{Range-based (RV).}
Since range images are 2D representations like RGB images, range-based 3D object detectors can naturally borrow the models in 2D object detection to handle range images \cite{meyer2019lasernet,bai2024rangeperception}.

Both of the above methods convert point clouds into a structured representation, with the main advantage being the ability to generate more dense feature maps. Even the space not occupied by the original point cloud learns features, which is crucial for sparse and incomplete objects. However, these methods inevitably result in information loss during the process of structuring the point cloud, and their reliance on large parameter 3D backbones makes it challenging to strike a balance between inference efficiency and detection accuracy. In contrast, in LiDAR scenes with multi-beams and fewer points, the advantages of structured point cloud detectors become less pronounced, and point-based detectors begin to stand out.

\begin{figure*}[t]
	\begin{center}
		\includegraphics[width=.7\textwidth]{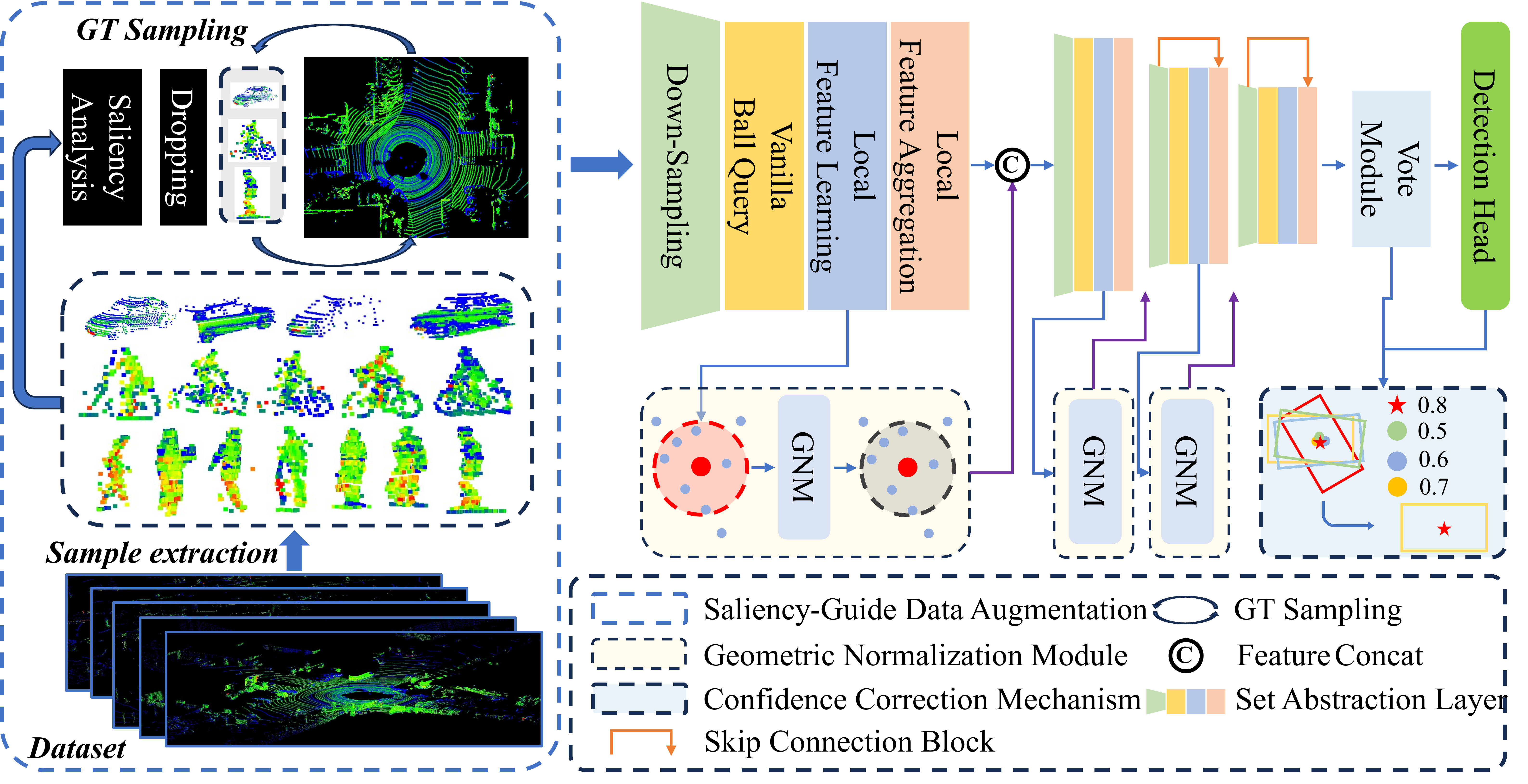}
	\end{center}
	\vspace{-0.3cm}
	\caption{Overview of proposed SGCCNet. SGCCNet adopts a PointNet++-style 3D backbone to learn point cloud features. In addition, SGCCNet consists of three core components, namely a saliency-guided data augmentation strategy, SA layer with geometric normalization modules and skip connection blocks, and a confidence correction mechanism during post-processing.}
	\label{fig3}
	\vspace{-0.3cm}
\end{figure*}

\textbf{Point-based.}
The current state-of-the-art point-based detectors \cite{yang20203dssd, chen2022sasa, huang2023object, zhang2022not,huang2023object, liang2023spsnet} follow the design paradigm of PointNet \cite{qi2017pointnet}, PointNet++ \cite{qi2017pointnet++}, and PointMLP \cite{ma2021rethinking}. The point cloud is processed through a set abstraction (SA) layer consisting of multiple stages of downsampling modules, local feature learning modules, and feature aggregation modules to learn rich spatial and semantic information. 

In terms of results, single-stage point-based detectors have achieved performance comparable to structure-based models in multi-beam LiDAR scenes, and their real-time inference capability and lightweight model architecture are desirable for many mobile embedded devices. The main performance bottlenecks are two factors: inadequate learning of low-quality targets (ILQ) and misalignment between localization accuracy and classification confidence (MLC).

\subsection{Dealing with ILQ}
As mentioned above, limited training data cannot cover all possible feature distributions. For models that rely on salient features, they cannot actively explore information in non-significant parts of the target. Inspired by AlexNet \cite{krizhevsky2012imagenet}, all current 3D detection models adopt some common data augmentation techniques to improve the feature distribution of training data, such as using geometric transformations to perturb both the overall scene and local features of the target. Choi et al. \cite{choi2021part} randomly occluded parts of the foreground target to expand the training data by 2.5 times. Reuse et al. \cite{reuse2021ambiguity} conducted detailed controlled experiments to demonstrate the effectiveness of local target transformations in improving detector performance. Hu et al. \cite{hu2021pattern} proposed pattern-aware GT sampling, enhancing data augmentation by subsampling objects based on LiDAR characteristics. Wang et al. \cite{wang2021pointaugmenting} consistently pasted virtual high-quality targets into point clouds to enhance the model's perception of low-quality targets.

These methods aim to improve model robustness by enhancing the diversity of the training data distribution. However, these augmentation processes are random, and data diversity does not necessarily imply feature diversity, leading to significant bottlenecks in benefits. This paper proposes a saliency-guided data augmentation method that, from the perspective of feature learning, removes salient features on which the model relies, forcing the model to actively explore information in low saliency regions and enhance the diversity of feature distribution in training samples from the source.

\subsection{Dealing with MLC}
He et al. \cite{9157660} proposed an auxiliary network to convert the features of key points belonging to the grid in the 3D backbone into point-wise representation to improve localization accuracy. They also introduced a part-sensitive warping operation to align the confidences to the predicted bounding boxes. CIA-SSD \cite{zheng2021cia} combined the predicted IoU with the key point classification probability as the final confidence. Wang et al. \cite{wang20213dioumatch} proposed a confidence-based filtering mechanism to filter out poorly localized proposals. Sheng et al. \cite{sheng2022rethinking} proposed a new Rotation-Decoupled IoU (RDIoU) method to generate more effective optimization targets. In addition, models such as FVTP  \cite{li2021voxel}, DFDNet \cite{liu2019deep}, and DDIGNet \cite{ming2023deep} have also adopted IoU prediction branches to improve the MLC problem.

Although the above methods have achieved certain effects, they are mostly designed for anchor-based detection heads, and point-based detection heads have not been given much attention. Specifically, compared to anchor-based detection heads, the sparsity of keypoints obtained by point-based detection heads varies, leading to different numbers of predicted bounding boxes for each target, making these methods unsuitable for direct use in point-based detection heads. In this paper, a new confidence correction mechanism is proposed for point-based detection heads, taking into account the prediction situations of other proposals within the keypoint domain and the density of key points in the domain. On the one hand, it can filter out low-quality proposals, and on the other hand, it can explore potential high-quality proposals.

\section{Method}
\label{sec:3}
\subsection{Overview}
As mentioned above, our goal is to improve the model's ability to learn from low-quality targets and enhance target localization accuracy by correcting confidence. To achieve this, we propose SGCCNet, a generic and unified single-stage point-based 3D object detector as illustrated in Fig. \ref{fig3}, which includes three core components, namely a saliency-guided data augmentation method, SA layers with geometric normalization modules and skip connection blocks, and a confidence correction mechanism.

In this section, we will detail the design principles and implementation process of SGCCNet. Section \ref{sec:3.2} introduces the mathematical symbols and their meanings to be used. In Section \ref{sec:3.3}, we present the method for obtaining point saliency in the saliency-guide data augmentation and the specific enhancement process. Section \ref{sec:34} describes the specific structure of the 3D backbone of SGCCNet, focusing on the design rationale of the geometric correction module and skip connection block. Section \ref{sec:3.5} proposes mechanisms for point-based detection heads and confidence correction. Finally, in Section \ref{sec:3.6}, we discuss the model loss in the end-to-end training process.
\subsection{Preliminary}
\label{sec:3.2}
Let $\mathcal{D}$ be the training dataset used for experiments, consisting of $N$ point cloud scenes $\mathcal{P}_i$, $\{\mathcal{P}_i \in \mathbb{R}^{n_i \times (3+C)} | i=1,2,...,N\}$. Here, $n_i$ represents the number of point clouds in the current scene, and $C$ represents other features besides the spatial positions of the point clouds, such as intensity. The annotated ground-truth for each scene is denoted as $\mathcal{B}_i$, $\{\mathcal{B}_i \in \mathbb{R}^{m_i \times 8} | i=1,2,...,N\}$, where $m_i$ is the number of ground-truth annotations in the scene. The eight-dimensional features include the center position of the bounding box $\{x,y,z\}$, dimensions $\{l,w,h\}$, and category $c$. The total number of annotated ground-truth in the dataset is $A=\sum_{i=1}^N{m_i}$.

\subsection{Saliency-Guided Data Augmentation (SGDA)}
\label{sec:3.3}
For LiDAR-based 3D object detection, limited training samples cannot cover all possible target feature distributions. In a training mode with limited data and repeated sampling, the model gradually relies on prominent features with common distributions and uses them as decision criteria without exploring features in other less prominent regions of the target. Liang et al. \cite{liang2023spsnet} have demonstrated the model's dependency on prominent features in their work SPSNet. SPSNet assigns a Gaussian soft label to foreground points and regresses geometric information belonging to Dirac bounding boxes, obtaining the saliency score of points through stability and perturbation adversities. Ultimately, with the scenario of discarding only 10 foreground points, the model's average AP drops rapidly by 8.61\%. This dependency on salient features is detrimental to the robustness of the model's detection.

\begin{figure*}[t]
	\begin{center}
		\includegraphics[width=.7\textwidth]{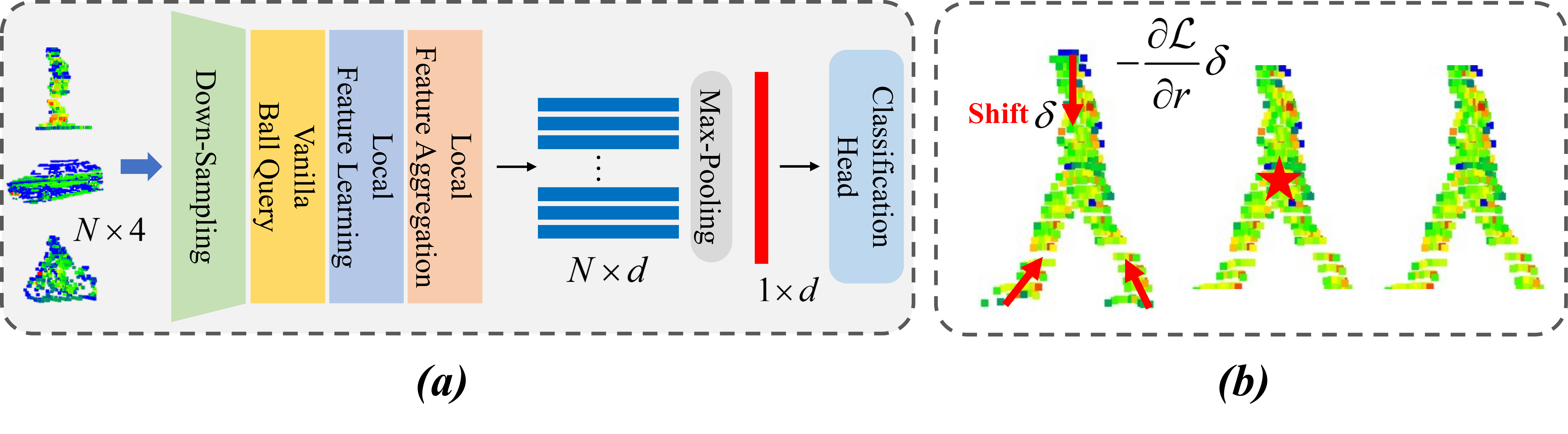}
	\end{center}
	\vspace{-0.3cm}
	\caption{(a) Overview of proposed SGCCNet-elite for classification task. (b) Shifting the point towards the centroid is similar to discarding the point, and the movement process is differentiable, which can be used to approximate the saliency score of the point.}
	\label{fig4}
	\vspace{-0.3cm}
\end{figure*}

We hope to develop a purposeful data augmentation approach, that is, from the perspective of features, by using saliency analysis to identify common features that the model relies on, removing them, and forcing the model to explore features in the target that were originally in low saliency regions. 
However, currently, there is not much attention given to point-wise saliency analysis methods for point-based 3D object detection. We have shifted our focus to developing more in-depth saliency analysis methods in point cloud classification tasks. Although these are two different tasks, we believe that the saliency scores obtained have commonalities for two reasons: 1) Considering the model structure. Point-based detectors throw vote points into the SA layer to determine the semantic information of local regions, a process that is identical to the prediction process of point cloud classification models. 2) Considering feature learning. Saliency scores are determined by the specific features of the model, and using similar structured models, the saliency features they focus on should be similar as well. We will demonstrate our viewpoint in the experimental section, and next we will introduce the specific data augmentation process.

\textbf{Dataset Preparing.}
We conduct our experiments on the KITTI dataset $\mathcal{D}$. Firstly, we extract all foreground objects using the annotated ground-truth $\{\mathcal{B}_i \in \mathbb{R}^{m_i \times 8} | i=1,2,...,A\}$ in $\mathcal{D}$, and then construct a classification dataset similar to ModelNet40 \cite{wu20153d} and ScanObjectNN \cite{uy2019revisiting}. Each point in the dataset contains 4-dimensional features, representing spatial position and intensity. KITTI categorizes foreground objects into \textit{Easy}, \textit{Moderate}, and \textit{Hard} levels based on factors such as occlusion and the number of points within the bounding box. We discard samples with fewer points than a certain threshold, and train the classification model on all levels. It is worth noting that we only perform data augmentation on \textit{Easy} and \textit{Moderate} levels, while retaining all samples in the \textit{Hard} level.

\textbf{Classification Model Preparing.}
We have streamlined the structure of SGCCNet and constructed an elite model SGCCNet-elite for classification tasks as shown in Fig. \ref{fig4} (a). This model consists of only one SA layer for feature learning, and the training samples do not undergo downsampling in the SA layer. Finally, the point-wise features of each point are pooled using max pooling to form the overall feature of the sample, which is then fed into a Fully Connected (FC) Layer for classification.

\textbf{Saliency Analysis.}
Let $E_{\mathcal{\theta}}(p^i)$ be a classification model, where $p^i:=(p^i_{1},p^i_{2},...,p^i_{k_i})$ is a training sample, and $p\in \mathbb{R}^{k_i\times 4}$. $\mathcal{L}_c(\cdot)$ is the classification loss function.
\begin{equation}
	\mathcal{L}_c=-\frac{1}{A}\sum_i{\sum_{c=1}^3{y_{ic}\log(\hat{y_{ic}})}}
\end{equation}
Among them, $y_{ic}$ is the label, with a value of 1 if sample $p^i$ belongs to class $c$, otherwise 0, and $\hat{y_{ic}}$ is the predicted probability. The most intuitive way to determine point significance is to gradually discard points and observe the change in loss $\mathcal{L}_c$. However, due to the large number of points and the fact that the contribution of each point to the whole is not isolated, this method is time-consuming and inaccurate. Inspired by Zheng et al. \cite{zheng2019pointcloud}, we transform the process of discarding points into a differentiable process of moving points to calculate the significance score of each point. Specifically, in the classification task, we normalize the feature information of each sample.
\begin{equation}
	\label{eq2}
	\hat{p^i} = \frac{p^i-{\frac{1}{k} \sum_{j=1}^{k_i}{p^i_{j}}}}{\max \limits_{j\in \{1,2,...,k_i\}} {\sqrt{\sum_{c=1}^{4}{p^i_{{jc}}}^2}}}
\end{equation}
The center of the sample is now moved to the origin. Based on the mechanism of LiDAR scanning imaging, we have reason to believe that the points at the origin have little contribution to the classification of the sample. The impact of moving points from other positions in the sample to the origin on model decisions is almost the same as discarding points. Since points are not angle invariant, there are difficulties in measuring gradients in Euclidean space, so we consider point shifting in the Spherical Coordinate System.

In the Spherical Coordinate System, a point ${p^i_{j}}$ is represented as $\left( {{r}^i_{j}},{{\varphi }^i_{j}},{{\phi }^i_{j}} \right)$ with $a$ as the sphere core, ${{r}^i_{j}}$ is distance of ${p^i_{j}}$ to $a$, ${{\varphi }^i_{j}}$ and ${{\phi }^i_{j}}$ are the two angles of a point relative to $a$. After shifting ${p^i_{j}}$ in the direction of ${{r}^i_{j}}$ towards sphere core $a$ by $\delta $, the change in model loss is $-\delta \frac{\partial \mathcal{L}_c}{\partial {{r}^i_{j}}}$, where ${{r}^i_{j}}=\sqrt{\underset{c=1}{\overset{3}{\mathop \sum }}\,{{\left( {{p}^i_{jc}}-{{a}_{c}} \right)}^{2}}}$, as: 

\begin{equation}
	\left\{
	\begin{matrix}
		{{p}^i_{j1}}-{{a}_{1}}={{r}^i_{j}}\cos {{\phi }^i_{j}}\sin {{\varphi }^i_{j}}  \\
		{{p}^i_{j2}}-{{a}_{2}}={{r}^i_{j}}\cos {{\varphi }^i_{j}}\sin {{\phi }^i_{j}}  \\
		{{p}^i_{j3}}-{{a}_{3}}={{r}^i_{j}}\sin {{\phi }^i_{j}}  \\
	\end{matrix}
	\right.
\end{equation}

where ${{p}^i_{j}}={{\left\{ {{p}^i_{jc}} \right\}}_{c=1,2,3}}$ , $a={{\left\{ {{a}_{c}} \right\}}_{c=1,2,3}}$ represent the 3D coordinates value of ${p^i_{j}}$ and $a$. So:

\begin{equation}
	\frac{\partial \left( {{p}^i_{jc}}-{{a}_{c}} \right)}{\partial {{r}^i_{j}}}=\frac{{{p}^i_{jc}}-{{a}_{c}}}{{{r}^i_{j}}}
\end{equation}
\begin{equation}
	\frac{\partial \mathcal{L}}{\partial {{r}^i_{j}}}=\underset{c=1}{\overset{3}{\mathop \sum }}\,\frac{\partial \mathcal{L}_c}{\partial {{p}^i_{jc}}}\frac{{{p}^i_{jc}}-{{a}_{c}}}{{{r}^i_{j}}}
\end{equation}

So that after shifting ${p^i_{j}}$, the change in model loss is ${{s}^i_{j}}=-\frac{\partial \mathcal{L}_c}{\partial {{r}^i_{j}}}{{r}^i_{j}}$. In practical computation, we also set the fourth dimension intensity value of the central position point to 0, and add it as a spatial information to the gradient calculation.

\begin{algorithm}[t]
	\caption{Dropping Points Strategy}
	\label{Alg:alg1}
	\begin{algorithmic}[1]
		\REQUIRE ~~\\
		Selected training sample $p^i=\{p^i_j|j=1,2,...,k_i\}$, $p^i_j=(x^i_j,y^i_j,z^i_j,I^i_j)\in \mathbb{R}^4$. Category label $c_i$;\\
		Point cloud classification model $E_\mathcal{\theta}(\cdot)$ and loss function $\mathcal{L}_c(\cdot)$;\\
		Linear hyperparameter of the number of deleted points $\alpha, \beta$ and dropout interval $drop\_interval$;\\
		\ENSURE ~~\\
		$k_i > 0$;
		\STATE{
			Calculate the number of points that should be discarded: $d^i=\lfloor \alpha k_i + \beta \lfloor$;
		}
		\FOR{$l = 0, 1, \cdots, \lfloor \frac{d^i}{drop\_interval} \rfloor$}
		
		\STATE{
			Get normalized point cloud $\hat{p^i}$ with Eq. \ref{eq2};
		}
		\STATE{
			Calculate classification loss: $\mathcal{L}=\mathcal{L}_c(E_\mathcal{\theta}(\hat{p^i}),c_i)$;
		}
		\STATE{
			Calculate the saliency score for each point: ${{s}^i_{j}}=-\frac{\partial \mathcal{L}}{\partial {{r}^i_{j}}}{{r}^i_{j}}$;
		}
		\STATE{
			Keep the lowest $k_i - drop\_interval$ saliency score points $p^i_{dropped}$;
		}
		\STATE{
			$k_i \gets k_i - drop\_interval$;
		}
		\STATE{
			$p^i \gets p^i_{dropped}$;
		}
		\ENDFOR
		\RETURN $p_i$
	\end{algorithmic}
\end{algorithm}

\begin{figure}[t]
	\begin{center}
		\includegraphics[width=3.5in]{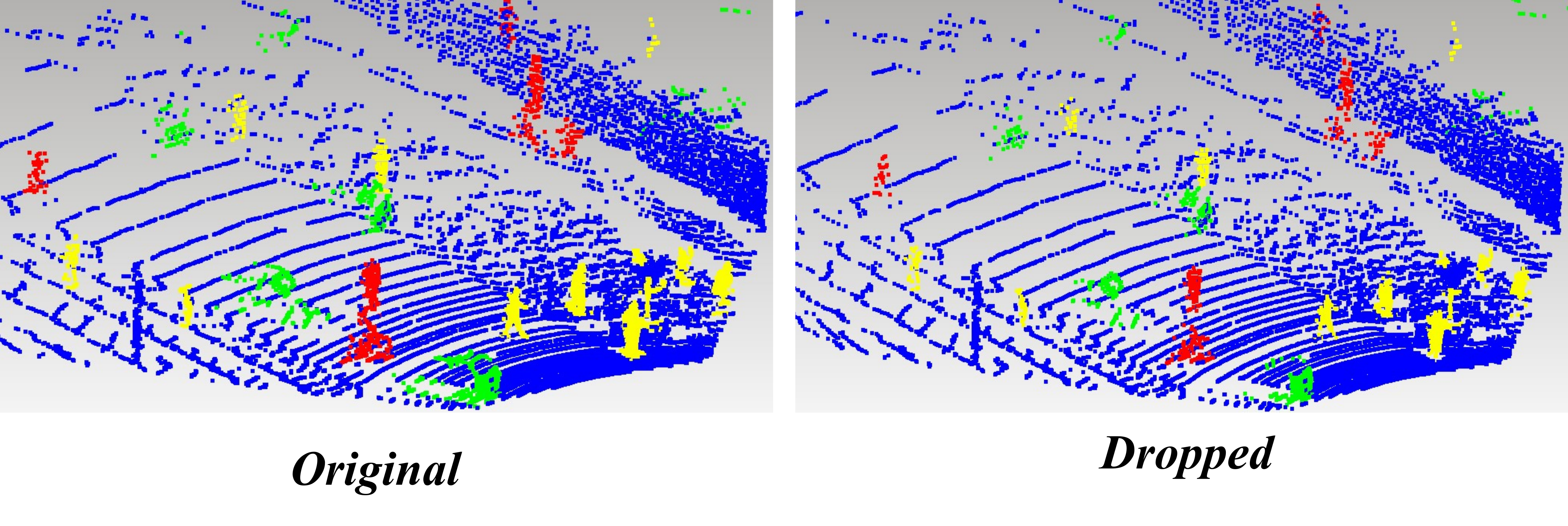}
	\end{center}
	\vspace{-0.3cm}
	\caption{Mark the changes in the scenes before and after dropping points, with points representing the \textit{Car}, \textit{Pedestrian}, \textit{Cyclist}, and \textit{Background} classes in green, yellow, red, and blue respectively.}
	\label{fig5}
	\vspace{-0.3cm}
\end{figure}

\begin{figure}[t]
	\begin{center}
		\includegraphics[width=3in]{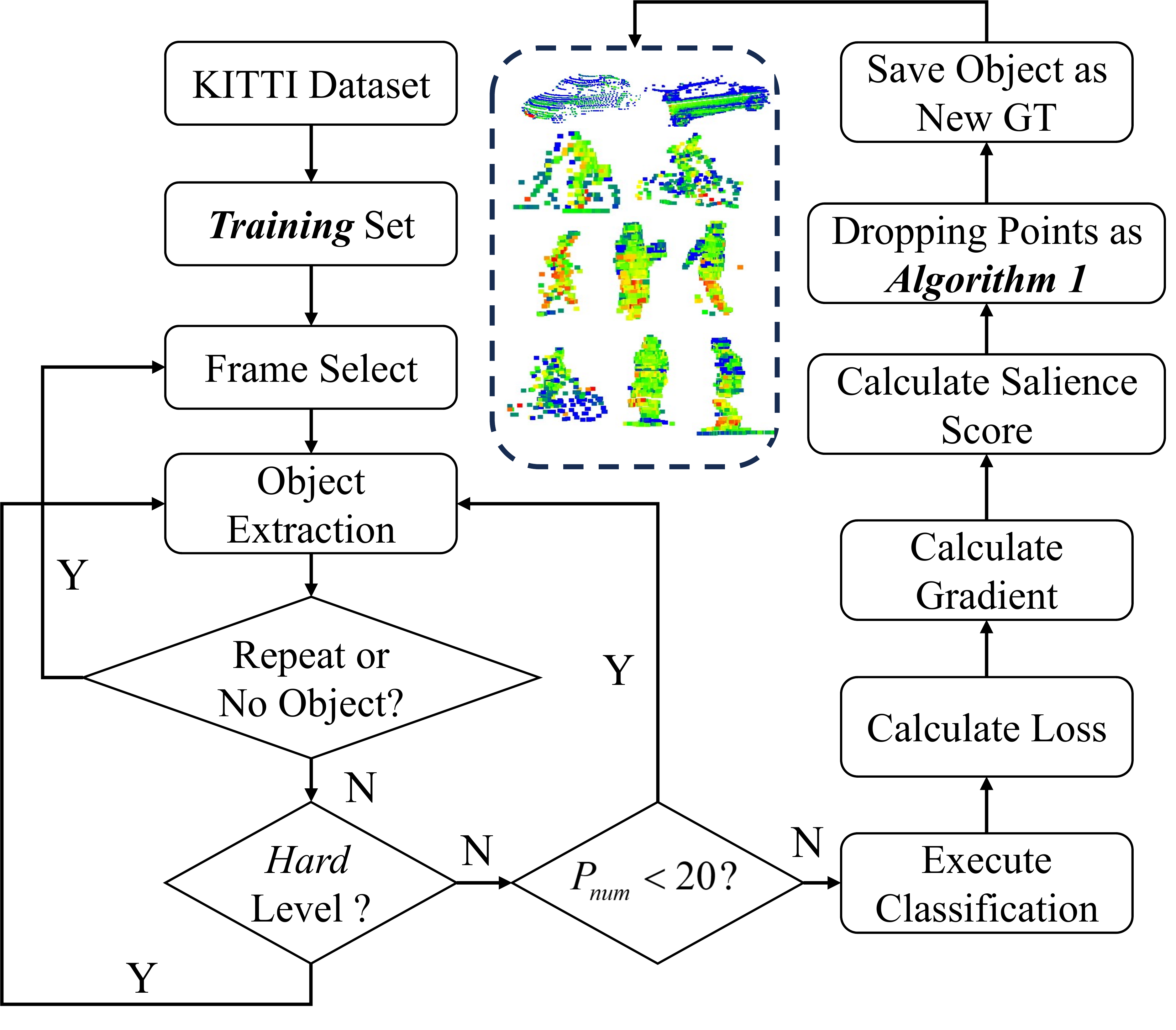}
	\end{center}
	\vspace{-0.3cm}
	\caption{Workflow of Saliency-Guided Data Augmentation (SGDA).}
	\label{fig6}
	\vspace{-0.3cm}
\end{figure}

\textbf{Droppint Points.}
For a training sample $p^i$, after obtaining the saliency scores $\{s^i_j|j=1,2,...,k_i\}$ for each point, we delete high saliency points in the sample according to a preset ratio. The number of points to be deleted, denoted as $d^i$, is linearly proportional to the number of points in the sample, $k$, i.e., $d^i=\lfloor \alpha k_i + \beta \rfloor$. To ensure the accuracy of saliency ranking, the $d^i$ points are not deleted all at once, but are deleted at fixed intervals as described in Algorithm. \ref{Alg:alg1}. After discarding salient features, the sample is restored to its original space and used for training alongside the original ground truth in the GT sampling process. Fig. \ref{fig5} illustrates the changes in a set of scenarios before and after point removal. The entire saliency-guided data augmentation process is shown in Fig. \ref{fig6}.

\subsection{3D Backbone}
\label{sec:34}

Similar to previous state-of-the-art point-based detectors, SSGCNet also utilizes a PointNet++-style 3D backbone to extract features from point clouds. This backbone incorporates multi-stage downsampling, multi-scale feature learning, and local feature aggregation processes to extract point-wise features with rich semantic and geometric information.

\begin{figure*}[t]
	\begin{center}
		\includegraphics[width=.7\textwidth]{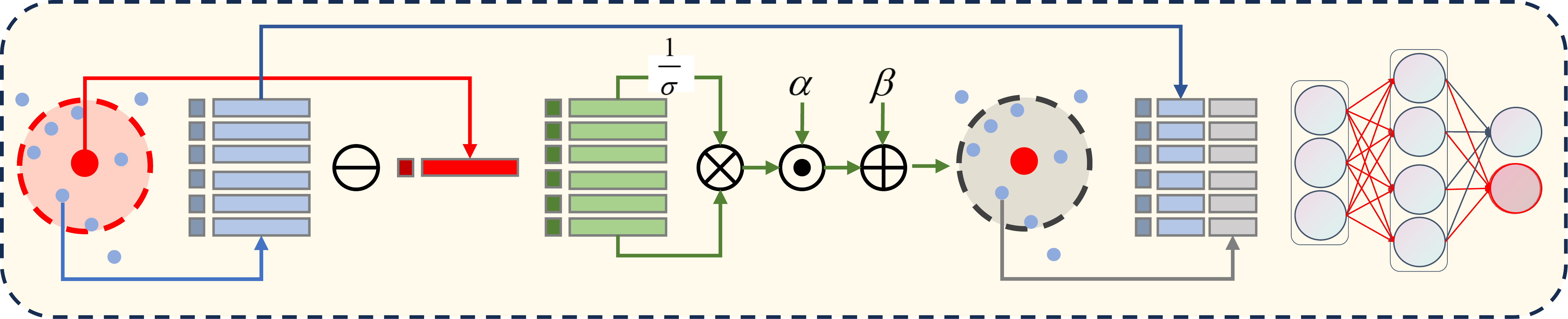}
	\end{center}
	\vspace{-0.3cm}
	\caption{Structure of the Geometric Normalization Module (GNM).}
	\label{fig7}
	\vspace{-0.3cm}
\end{figure*}

It is worth noting that the downsampling scheme of SGCCNet is designed based on IA-SSD. Due to the limited accuracy of semantic information learning by the point-based backbone, SGCCNet uses the Farthest Point Sampling (FPS) algorithm in the first two stages, and in the last two stages, sampling is based on the predicted point-wise foreground probability. The former ensures global coverage of sampling points, while the latter ensures a high recall rate of foreground points, reducing target information loss. In addition, SGCCNet also includes a geometric normalization module and skip connection block.

\textbf{Geometric normalization module (GNM).} 
We found that the PointNet structure ignores the Internal Covariate Shift (ICS) problem caused by the irregular and sparse geometric properties of point clouds during the process of aggregating local features. Specifically, in each learning stage, the same MLP layer needs to deal with a large number of regions with different geometric information simultaneously, and it is difficult for a simple MLP to achieve stable convergence speed and generalization ability from such a diverse feature distribution. Therefore, we propose to alleviate this problem by performing geometric normalization on the locally aggregated features, treating each ball used for learning local features as a batch, and applying 'Batch Normalization (BN)' operation before inputting into the MLP layer. The structure of GNM is shown in Fig. \ref{fig7}. In detail, we first move the distribution of local features within each ball to a more normalized space with the ball center as the mean, and then increase the diversity of the feature distribution through learnable weight and offset parameters of the same dimension as the features. Finally, to prevent information loss, we concat the normalized features with the original features and input them into the MLP layer for learning local features.

The specific structure of the geometric normalization module is shown in Fig. \ref{fig7}. Let $\{f_{i,j}\}_{j=1,2,...,k}\in \mathbb{R}^{k\times d}$ be the local features of the sampling point $p_i$, where $k$ is the number of neighbors in its local region, and $d$ is the feature dimension of the sampling point and its neighbor points. We normalize the features of neighbor points in the local region through the following equation:
\begin{equation}
	\{f_{i,j}\}=\alpha \odot \frac{ \{f_{i,j}\}-f_i}{\sigma+\epsilon}+\beta
\end{equation}
\begin{equation}
	\sigma=\sqrt{\frac{1}{k\times n\times d}\sum_{i=1}^n \sum_{j=1}^k(f_{i,j}-f_i)^2}
\end{equation}
Similar to BN layer, where $\alpha \in \mathbb{R}^d$ and $\beta \in \mathbb{R}^d$ are learnable parameters, and $\odot$ indicates Hadamard product. $\epsilon=1e^{-5}$ is a small number for numerical stability. Note that $\sigma$ is a scalar that describes the feature deviation across all local groups and channels. By doing so, we transform the points via a normalization operation while maintaining original geometric properties. The features after geometric normalization will be concatenated with the original features for subsequent learning.

\begin{figure}[t]
	\begin{center}
		\includegraphics[width=3.5in]{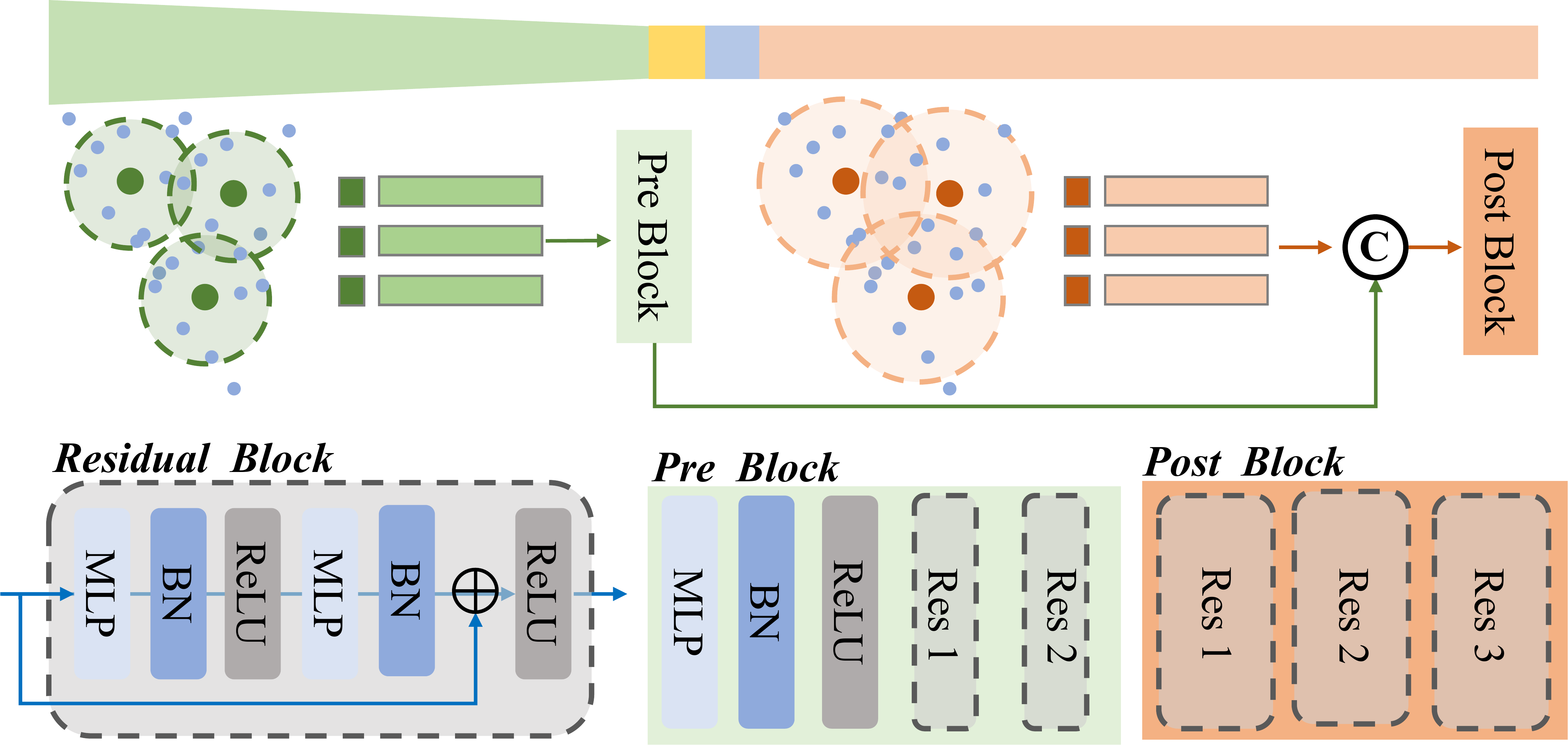}
	\end{center}
	\vspace{-0.3cm}
	\caption{Structure of the Skip Connection Block (SCB).}
	\label{fig8}
	\vspace{-0.3cm}
\end{figure}

\textbf{Skip Connection Block (SCB).}
When determining the appropriate number of training epochs for SGCCNet, we found that some high-quality targets were detected with high confidence early in training, but as training progressed, the confidence decreased below the threshold, undoubtedly reducing the model's performance. After analysis, we believe there may be three reasons for this: 1) The sparsity of the targets greatly increases after continuous downsampling, causing sparse sampling points to lose the ability to perceive the overall geometric information of the targets. 2) As the ball query radius increases, noise and features of other targets in the local area are introduced while expanding the model's receptive field, leading to the loss of key information of the corresponding targets after pooling layers. 3) For some targets, shallow features have a better ability to represent target information. In testing classification tasks, the fact that models with extremely small parameter amounts can achieve classification levels far higher than detection models directly proves this point. We refer to this type of problem as feature forgetting and have designed a skip connection block to alleviate this issue. The structure of SCB is shown in Fig. \ref{fig8}.

The SCB alleviates the problem of feature forgetting by enabling interactions between key point features of adjacent SA layers in the 3D backbone, as shown in Fig. \ref{fig8}. Let the feature of the $l$-th SA layer's sampled points be $\{F^{l}_{i}\}_{i=1,...,k}\in \mathbb{R}^{k\times d_l}$, where $k$ is the number of sampled points at the current stage, and $d$ is the feature dimension. Based on the basic structure of PointNet++, it is known that the sampled points at the current stage are also sampled points from the previous stage. Therefore, let the features of the sampled points in the $l$-th SA layer at the previous stage be $\{F^{l-1}_{i}\}_{i=1,...,k}\in \mathbb{R}^{k\times d_{l-1}}$. The learning process of SCB is as follows: (insert equation here).
\begin{equation}
	g_i=\Phi_{pos}(\Phi_{pre}(F^{l-1}_{i})+F^{l}_{i})
\end{equation}
The $\Phi_{pre}$ is a module composed of residual networks, which elevates the dimension of features $F^{l-1}_{i}$ from $d_{l-1}$ to $d_l$ from the previous stage, and $\Phi_{pos}$ further interacts with the features. The design of SCB is inspired by He et al.'s work \cite{9157660}, but we naturally leverage the advantages of point-based backbone hierarchical downsampling, without any feature projection or transformation processes. As shown in Fig. \ref{fig8}, with the addition of SCB, shallow features and deep features of the same sampling point, as well as local small region features and large region features, interact with each other. Even in the subsequent downsampling process where the sampling points become sparser or noise is introduced, valuable information from earlier stages is still retained.
\begin{algorithm}[!t]
	\caption{Confidence Correction Mechanism}
	\label{Alg:alg2}
	\begin{algorithmic}[1]
		\REQUIRE ~~\\
		Predicted bounding boxes $\mathcal{B}$ of one LiDAR frame with the size of $k \times 7$, where $k$ is the number of bounding boxes, and $(x, y, z, w, l, h, r)$ is the parameters of a bounding box;\\
		Predicted classification confidence values $\mathcal{C}$ and IoU values $\mathcal{U}$ of the corresponding predicted bounding boxes with the size of $N \times 1$, respectively;\\
		Initial confidence score threshold $score\_thres_1$. Final confidence score threshold $score\_thres_2$;\\
		IoU threshold $iou\_thres$;\\
		Missed sample incremental confidence $\Delta c$. Missed sample IoU threshold $iou\_thres_{missed}$. Threshold of the number of missed sample neighbors $neighbor\_thres_{missed}$;\\
		$\mathcal{B} = \{\mathbf{b}_i\}_{i=1,...,k}$; $\mathcal{C} = \{\mathbf{c}_i\}_{i=1,...,k}$; $\mathcal{U} = \{\mathbf{u}_i\}_{i=1,...,k};$\\
		\ENSURE ~~\\
		Initial selected bounding boxes index $\mathcal{D} = \emptyset$;\\
		Selected bounding boxes $\mathcal{B}_o = \emptyset$;\\
		Rectified confidence values $\mathcal{S}_o = \emptyset$ of the corresponding selected bounding boxes;
		\FOR{$i = 0, 1, \cdots, k$}
		\STATE{$k_{select} = 0$;}
		\IF{$\textbf{c}_i > score\_thres_1$}
		\STATE{$\textbf{c}_i \gets \textbf{c}_i^{0.7}\times \textbf{u}_i^{0.3}$;}
		\STATE{$k_{select} \gets k_{select} + 1$;}
		\STATE{$\mathcal{D} \gets \mathcal{D} \cup i$;}
		\ENDIF
		\ENDFOR
		\STATE{$\mathcal{B} \gets \mathcal{B}(\mathcal{D})$; $\mathcal{C} \gets \mathcal{C}(\mathcal{D})$; $\mathcal{U} \gets \mathcal{U}(\mathcal{D})$;}
		\FOR{$i = 0, 1, \cdots, k_{select}$}
		\STATE{$iou_{all} = 0, N_{neighbor} = 0$;}
		\FOR{$j = 0, 1, \cdots, k_{select}$}
		\IF{$\mathrm{IoU}(\mathbf{b}_i, \mathbf{b}_j) > iou\_thres$}
		\STATE{
			$iou_{all} \gets  iou_{all} + \mathrm{IoU}(\mathbf{b}_i, \mathbf{b}_j)$;
		}
		\STATE{
			$N_{neighbor} \gets  N_{neighbor} + 1$;
		}
		\ENDIF
		\ENDFOR
		\STATE{
			$iou_{mean} = \frac{iou_{all}}{N_{neighbor}};$
		}
		\STATE{
			$s = iou_{mean} \cdot c_i;$
		}
		\IF{$iou_{mean} > iou\_thres_{missed} \quad \& \quad N_{neighbor} > neighbor\_thres_{missed}$}
		\STATE{
			$s \gets s + \Delta c;$
		}
		\ENDIF
		\IF{$s > score\_thres_2$}
		\STATE{
			$\mathcal{B}_o \gets \mathcal{B}_o \cup \mathbf{b}_i$;\\
			$\mathcal{S}_o \gets \mathcal{S}_o \cup s$;
		}
		\ENDIF
		\ENDFOR
		\RETURN $\mathcal{B}_o, \mathcal{S}_o$
	\end{algorithmic}
\end{algorithm}

\subsection{Confidence Correction Mechanism (CCM)}
\label{sec:3.5}

The previous advanced single-stage point-based detectors use two completely independent branch networks in the detection head to respectively regress the confidence of predicted boxes and geometric information. During post-processing, the NMS operator is guided by confidence scores, retaining predicted boxes with the highest confidence scores in local regions and at any position with confidence scores higher than a threshold. This is disadvantageous for single-stage point-based detectors, as high confidence scores do not necessarily represent high-quality localization accuracy, and the rich geometric information of point clouds, coupled with the lack of semantic information, makes it difficult for the point-based backbone to learn precise confidence, leading to a large number of missed detections and false positive detections. Researchers have referred to this issue as misalignment between localization accuracy and classification confidence.

As mentioned in Section 2, the method with IoU prediction branch brings hope to alleviate the MLC problem of single-stage detectors. These methods predict the confidence of proposals while also predicting their IoU values with GT boxes. The final confidence sent to NMS is a joint indicator of both. SGCCNet also adopts this method to improve model performance. Furthermore, we not only consider the prediction of a single voting point but also incorporate the predictions of surrounding voting points into the reference range for confidence correction. The work closest to our idea is NIV-SSD \cite{liu2024niv}, but it only applies to single-class detection with anchor-based detection heads, and the issue of uneven prediction of each target voting point in point-based detectors is not considered.

\begin{figure}[t]
	\begin{center}
		\includegraphics[width=3.5in]{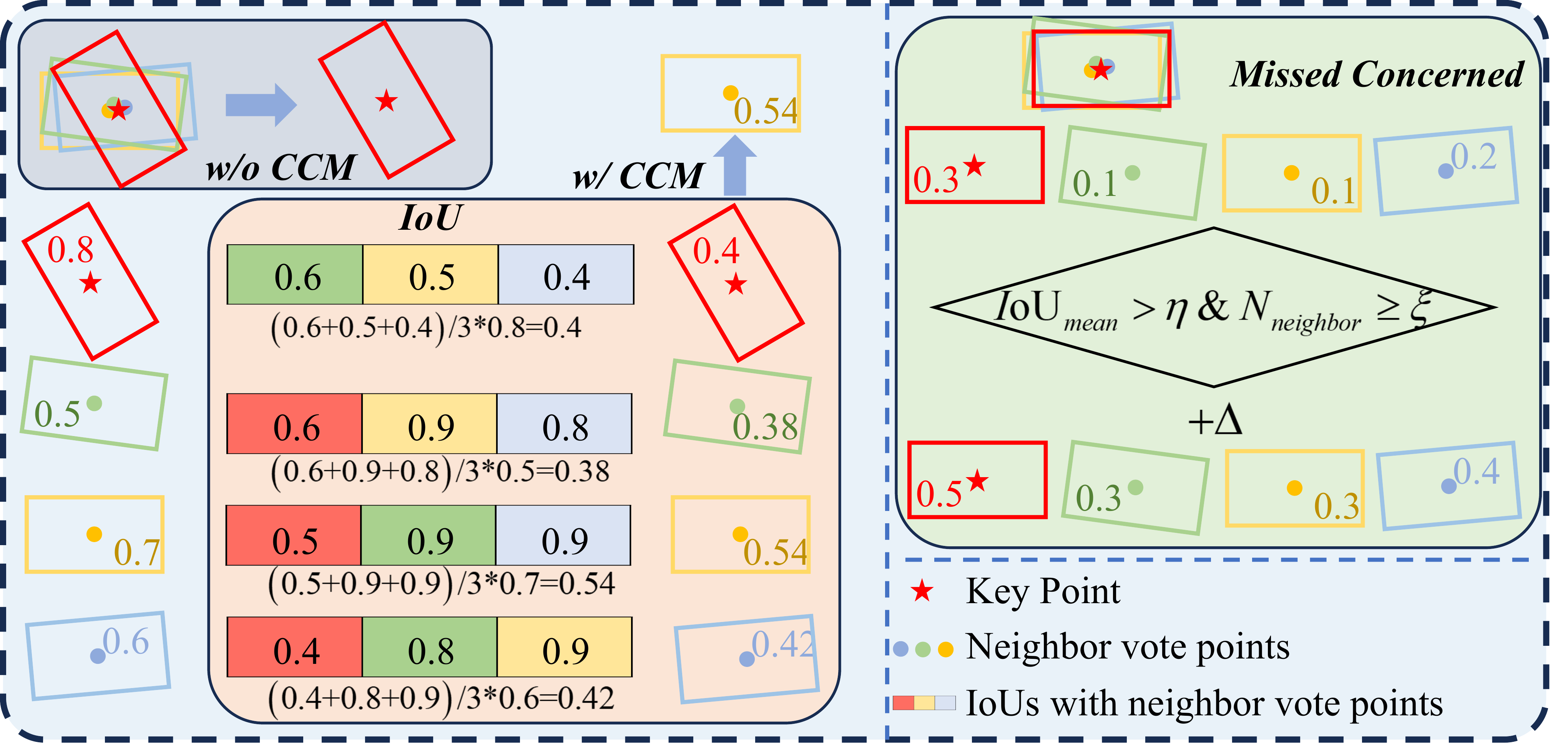}
	\end{center}
	\vspace{-0.3cm}
	\caption{The working mechanism of the confidence correction mechanism. CCM corrects the prediction of the current vote point by considering the predictions of other vote points in the neighborhood. CCM also takes into account the possibility of missed detections, where $\eta$ and $\xi$ in the figure correspond to the IoU threshold and the number of local neighbors threshold in line 20 of Algorithm \ref{Alg:alg2}.}
	\label{fig9}
	\vspace{-0.3cm}
\end{figure}

For a stable training SGCCNet, the key points output by the 3D backbone after voting will move to positions near the center of the target box. As shown in Fig. \ref{fig9}, for the same target, different voting points have a certain degree of clustering in terms of their positions and regression predicted boxes. Therefore, we use the predictions of neighboring voting points to correct the confidence of the current voting point. We believe that voting points in local regions with more neighbors having high IoU values are more accurate, while those with unreliable confidence. The specific confidence correction mechanism works as shown in Algorithm \ref{Alg:alg2}. Let the parameters of the predicted box regressed by the current voting point be $b_i=(x,y,z,w,l,h,r)$, we calculate its IoU value with the predicted boxes of other voting points $B=\{b_j\}_{j=1,...,k}\in \mathbb{R}^{k\times 7}$, and consider the predicted boxes with IoU values greater than a certain threshold as neighbors of the current voting point. Finally, the average IoU value between the voting point and its neighboring voting points' predicted boxes will be used as a weight to calibrate the confidence of the current voting point. In addition, CCM sets strict conditions to filter out boxes with high localization accuracy but confidence lower than a threshold. Before the NMS operation, we give them a certain increment to increase their probability of being considered as foreground. We will analyze the advantages of CCM in detail during the experimental phase.

\subsection{End-to-End Training}
\label{sec:3.6}

SGCCNet is a one-stage detector, using an end-to-end multi-task training fashion. Firstly, there is the semantic loss $\mathcal{L}_s$. The last two SA layers in SGCCNet use foreground sampling, so it is necessary to obtain the semantic information of the sampled points from the previous stage. Due to the highly unbalanced number of different categories of objects in the KITTI dataset, such as the number of \textit{Car} objects being five times that of \textit{Pedestrian} objects and ten times that of \textit{Cyclist} objects, and the point number of \textit{Car} objects is usually more than the other two, we adopt weighted cross entropy (WCE) loss to manually emphasize rare categories. The WCE loss can be formulated as:
\begin{equation}
	\alpha_c=\frac{1}{F_c-\epsilon}
\end{equation}
\begin{equation}
	\mathcal{L}_{s}=- \sum_{c=1}^C \alpha_c y_c \log{\hat{y_c}}
\end{equation}
where $y_c$ is the ground truth label determined by whether the point is inside the bounding box, $\hat{y_c}$ is the predicted probability, $F_c$ is the frequency, and $\alpha_c$ is the weight of the $c^{th}$ class. $C$ is the category number of the dataset. Yang et al. \cite{yang20203dssd} pointed out that points closer to the target center are easier to regress accurate bounding boxes, have higher confidence, and achieve higher accuracy in local semantic prediction. Following the design of Zhang et al. \cite{zhang2022not}, we assign weights to each point participating in the calculation of semantic loss, with points closer to the bounding box center having higher weights. This way, during training, these points will learn higher semantic scores.
\begin{equation}
	\mathcal{L}_{sample}=\sum_{i} \mathcal{L}_{{s}_i} \cdot MASK_i
\end{equation}
\begin{equation}
	Mask_i=\sqrt[3]{\frac{\min{f^*,b^*}}{\max{f^*,b^*}}\times \frac{\min{l^*,r^*}}{\max{l^*,r^*}} \times \frac{\min{u^*,d^*}}{\max{u^*,d^*}}}
\end{equation}
Where $Mask$ is the weight for the centrality of sampled points. $f^*,b^*,l^*,r^*,u^*,d^*$ represent the distances between the sampled points and the six faces of the target box. After training, points closer to the center of the target $\hat{y_c}$ will have higher values, and these points will be prioritized during inference.

The design of the remaining losses is similar to most point-based detectors. First is the regression loss of the vote points $\mathcal{L}_{vote}$, which determines the localization accuracy of the predicted boxes, we use $L1$ loss. Then is the bounding box classification prediction loss $\mathcal{L}_{cls}$, also calculated using WCE loss. Next is the regression loss of the vote point features for the bounding box sizes $\mathcal{L}_{reg}$, which is further decomposed into location, size, angle-bin, angle-res, and corner parts. To alleviate the MLC problem, we also add an IoU branch $\mathcal{L}_{IoU}$ to predict the IoU value between the predicted box regression of the vote points and the GT, also using $L1$ loss, the result of which is used to preliminarily correct the confidence (see Algorithm. \ref{Alg:alg2}). All the losses involved in the model are as follows:

\begin{equation}
	\mathcal{L}=\mathcal{L}_{sample}+\mathcal{L}_{vote}+\mathcal{L}_{cls}+\mathcal{L}_{reg}
\end{equation}
\begin{equation}
	\mathcal{L}_{reg}=\mathcal{L}_{loc}+\mathcal{L}_{size}+\mathcal{L}_{angle-bin}+\mathcal{L}_{angle-res}+\mathcal{L}_{corner}+\mathcal{L}_{IoU}
\end{equation}

\begin{table*}
	\renewcommand\arraystretch{.6}
	\centering
	\setlength{\extrarowheight}{0pt}
	\addtolength{\extrarowheight}{\aboverulesep}
	\addtolength{\extrarowheight}{\belowrulesep}
	\setlength{\aboverulesep}{0pt}
	\setlength{\belowrulesep}{0pt}
	\caption{Quantitative comparison with state-of-the-art methods on the KITTI \textit{test} set for \textit{Car} BEV and 3D detection, under the evaluation metric of 3D Average Precision ($AP$) of 40 sampling recall points. The best and our SGCCNet results are highlighted in \textbf{BOLD} and \uline{underlined}, respectively}
	\begin{tabular}{l|l|l|lll|lll} 
		\toprule
		\multirow{3}{*}{Method}                             & \multirow{3}{*}{Backbone} & \multirow{3}{*}{Type} & \multicolumn{3}{l|}{3D(IoU=0.7)} & \multicolumn{3}{l}{BEV(IoU=0.7)}  \\ 
		\cline{4-9}
		&                           &                       & \multicolumn{3}{l|}{R40}         & \multicolumn{3}{l}{R40}           \\
		&                           &                       & Easy    & Moderate & Hard        & Easy    & Moderate & Hard         \\ 
		\hline
		VoxelNet \cite{zhou2018voxelnet}                   & Voxel-based               & 1-stage               & $77.47$ & $65.11$  & $57.73$     & $87.95$ & $78.39$  & $71.29$      \\
		PointPillars \cite{lang2019pointpillars}                                        & Voxel-based               & 1-stage               & $82.58$ & $74.31$  & $68.99$     & $90.07$ & $86.56$  & $82.81$      \\
		SECOND \cite{s18103337}                                             & Voxel-based               & 1-stage               & $84.65$ & $75.96$  & $68.71$     & $89.39$ & $83.77$  & $78.59$      \\
		3DIoULoss \cite{zhou2019iou}                                          & Voxel-based               & 2-stage               & $86.16$ & $76.5$   & $71.39$     & $90.23$ & $86.61$  & $86.37$      \\
		TANet \cite{liu2020tanet}                                              & Voxel-based               & 1-stage               & $84.39$ & $75.94$  & $68.82$     & $91.58$ & $86.54$  & $81.19$      \\
		Part-A2 \cite{shi2020points}                                            & Voxel-based               & 2-stage               & $87.81$ & $78.49$  & $73.51$     & $91.7$  & $87.79$  & $84.61$      \\
		CIASSD \cite{zheng2021cia}                                             & Voxel-based               & 1-stage               & $89.59$ & $80.28$  & $72.87$     & $93.74$ & $89.84$  & $82.39$      \\
		SASSD \cite{he2020structure}                                              & Voxel-based               & 1-stage               & $88.75$ & $79.79$  & $74.61$     & $93.74$ & $89.84$  & $82.39$      \\
		Associate-3Det \cite{du2020associate}                                     & Voxel-based               & 1-stage               & $85.99$ & $77.4$   & $70.53$     & $91.4$  & $88.09$  & $82.96$      \\
		SAT-GCN \cite{wang2023sat}                                            & Voxel-based               & 1-stage               & $86.55$ & $78.12$  & $73.72$     & $92.83$ & $88.06$  & $83.51$      \\
		SVGA-Net \cite{he2022svga}                                           & Voxel-based               & 1-stage               & $87.33$ & $80.47$  & $75.91$     & -       & -        & -            \\ 
		\hline
		Fast Point R-CNN \cite{chen2019fast}                                   & Point-Voxel               & 2-stage               & $85.29$ & $77.4$   & $70.24$     & $90.87$ & $87.84$  & $80.52$      \\
		STD \cite{yang2019std}                                                & Point-Voxel               & 2-stage               & $87.92$ & $79.71$  & $75.09$     & -       & -        & -            \\
		PV-RCNN \cite{shi2020pv}                                            & Point-Voxel               & 2-stage               & $\uline{90.25}$ & $81.43$  & $76.82$     & $\uline{94.98}$ & $\uline{90.62}$  & $86.14$      \\
		EQ-PVRCNN \cite{yang2022unified}                                          & Point-Voxel               & 2-stage               & $90.13$ & $\uline{82.01}$  & $\uline{77.53}$     & $94.55$ & $89.09$  & $\uline{86.42}$      \\
		VIC-Net \cite{jiang2021vic}                                            & Point-Voxel               & 1-stage               & $88.25$ & $80.61$  & $75.83$     & -       & -        & -            \\
		HVPR \cite{noh2021hvpr}                                               & Point-Voxel               & 1-stage               & $86.38$ & $77.92$  & $73.04$     & -       & -        & -            \\ 
		\hline
		PointRCNN \cite{shi2019pointrcnn}                                          & Point-based               & 2-stage               & $86.96$ & $75.64$  & $70.7$      & $92.13$ & $87.39$  & $82.72$      \\
		3D IoU-Net \cite{li20203d}                                         & Point-based               & 2-stage               & $87.96$ & $79.03$  & $72.78$     & $94.76$ & $88.38$  & $81.93$      \\
		3DSSD$^\mathbf{\dag}$ \cite{yang20203dssd}                                              & Point-based               & 1-stage               & $88.36$ & $79.57$  & $74.55$     & $92.66$ & $89.02$  & $85.86$      \\
		IA-SSD \cite{zhang2022not}                                             & Point-based               & 1-stage               & $88.34$ & $80.13$  & $75.04$     & $92.79$ & $89.33$  & $84.35$      \\
		IA-SSD* \cite{zhang2022not}                                            & Point-based               & 1-stage               & $87.14$ & $78.47$  & $73.53$     & $92.21$ & $88.71$  & $83.72$      \\
		DBQ-SSD \cite{yang2022dbq}                                            & Point-based               & 1-stage               & $87.93$ & $79.39$  & $74.4$      & -       & -        & -            \\
		 SGCCNet            & Point-based               & 1-stage               & $\textbf{89.24}$ & $\textbf{80.82}$  & $\textbf{75.58}$     & $\textbf{93.57}$ & $\textbf{89.88}$  & $\textbf{85.27}$      \\
		 \textit{-Fast Point R-CNN} & -                         & -     & $\textit{3.95}$  & $\textit{3.42}$   & $\textit{5.34}$      & $\textit{2.7}$   & $\textit{2.04}$   & $\textit{4.75}$       \\
		 \textit{-IA-SSD}           & -                         & -                     & $\textit{0.9}$   & $\textit{0.69}$   & $\textit{0.54}$      & $\textit{0.78}$  & $\textit{0.55}$   & $\textit{0.92}$       \\
		 \textit{- IA-SSD*}          & -                         & -                     & $\textit{2.1}$   & $\textit{2.35}$   & $\textit{2.05}$      & $\textit{1.36}$  & $\textit{1.17}$   & $\textit{1.55}$       \\
		\bottomrule
	\end{tabular}
	\label{tab:1}
\end{table*}

\section{Experiments}
\label{sec:4}
In this section, we will provide detailed experiments to demonstrate the efficiency and accuracy of SGCCNet. Specifically, we introduced the specific settings and implementation details of the experiments in Section \ref{sec:4.1}. Then, an analysis of SGCCNet's detection performance on the KITTI dataset and a comparison with other state-of-the-art models is reported in Section \ref{sec:4.2}. In Section \ref{sec:4.3}, we analyze the inference efficiency of SGCCNet. Furthermore, various ablation experiments are conducted in Section \ref{sec:44} to demonstrate the effectiveness of the SGCCNet design. Finally, in Section \ref{sec:45}, we add our model components to other advanced models to demonstrate the effectiveness of our design.

\begin{table*}
	\renewcommand\arraystretch{.6}
	\label{tab:2}
	\centering
	\setlength{\extrarowheight}{0pt}
	\addtolength{\extrarowheight}{\aboverulesep}
	\addtolength{\extrarowheight}{\belowrulesep}
	\setlength{\aboverulesep}{0pt}
	\setlength{\belowrulesep}{0pt}
	\caption{Quantitative comparison with state-of-the-art methods on the KITTI \textit{val} set for \textit{Car} BEV and 3D detection, under the evaluation metric of 3D Average Precision ($AP$) of 40 sampling recall points. The best and our SGCCNet results are highlighted in \textbf{Bold} and \uline{underlined}, respectively}
	\resizebox{\textwidth}{!}{
	\begin{tabular}{l|l|l|lll|lll|lll|lll} 
		\toprule
		\multirow{3}{*}{Method}                   & \multirow{3}{*}{Backbone} & \multirow{3}{*}{Type} & \multicolumn{6}{l|}{3D(IoU=0.7)}                & \multicolumn{6}{l}{BEV(IoU=0.7)}                  \\ 
		\cline{4-15}
		&                           &                       & \multicolumn{3}{l|}{R11}     & \multicolumn{3}{l|}{R40}     & \multicolumn{3}{l|}{R11}     & \multicolumn{3}{l}{R40}       \\
		&                           &                       & Easy    & Moderate & Hard    & Easy    & Moderate & Hard    & Easy    & Moderate & Hard    & Easy    & Moderate & Hard     \\ 
		\hline
		VoxelNet \cite{zhou2018voxelnet}                                 & Voxel-based               & 1-stage               & $81.97$ & $65.46$  & $62.85$ & -       & -        & -       & $89.6$  & $84.81$  & $78.57$ & -       & -        & -        \\
		PointPillars \cite{lang2019pointpillars}                             & Voxel-based               & 1-stage               & $86.44$ & $77.28$  & $74.65$ & $87.75$ & $78.38$  & $75.18$ & $89.66$ & $87.16$  & $84.39$ & $92.05$ & $88.05$  & $86.67$  \\
		SECOND \cite{s18103337}                                   & Voxel-based               & 1-stage               & $88.61$ & $78.62$  & $77.22$ & $90.55$ & $81.61$  & $78.61$ & $90.01$ & $87.92$  & $86.45$ & $92.42$ & $88.55$  & $87.65$  \\
		SECOND-iou \cite{s18103337}                               & Voxel-based               & 1-stage               & $84.93$ & $76.3$   & $75.98$ & $86.77$ & $79.23$  & $77.17$ & $87.9$  & $76.3$   & $75.95$ & $90.23$ & $86.61$  & $86.37$  \\
		TANet \cite{liu2020tanet}                                    & Voxel-based               & 1-stage               & $88.17$ & $77.75$  & $75.31$ & -       & -        & -       & -       & -        & -       & -       & -        & -        \\
		Part-A2 \cite{shi2020points}                                  & Voxel-based               & 2-stage               & $89.55$ & $79.41$  & $78.85$ & $92.15$ & $82.91$  & $82.05$ & $90.2$  & $87.96$  & $87.56$ & $92.9$  & $90.01$  & $88.35$  \\
		Part-A2-free \cite{shi2020points}                             & Voxel-based               & 1-stage               & $89.12$ & $78.73$  & $77.98$ & $91.68$ & $80.31$  & $78.1$  & $90.1$  & $86.79$  & $84.6$  & $92.84$ & $88.15$  & $86.16$  \\
		SASSD \cite{he2020structure}                                    & Voxel-based               & 1-stage               & $89.69$ & $79.41$  & $78.33$ & -       & -        & -       & $\uline{90.59}$ & $88.43$  & $87.49$ & -       & -        & -        \\
		Associate-3Det \cite{du2020associate}                           & Voxel-based               & 1-stage               & $0$     & $79.17$  & -       & -       & -        & -       & -       & -        & -       & -       & -        & -        \\
		CIASSD \cite{zheng2021cia}                                   & Voxel-based               & 1-stage               & $0$     & $79.81$  & -       & -       & -        & -       & -       & -        & -       & -       & -        & -        \\ 
		\hline
		PV-RCNN \cite{shi2020pv}                                  & Point-Voxel               & 2-stage               & $89.26$ & $79.16$  & $\uline{79.39}$ & $91.37$ & $82.78$  & $80.24$ & $89.98$ & $87.7$   & $86.59$ & $92.72$ & $88.59$  & $88.04$  \\
		Fast Point R-CNN \cite{chen2019fast}                         & Point-Voxel               & 2-stage               & $0$     & $79$     & -       & -       & -        & -       & -       & -        & -       & -       & -        & -        \\
		STD \cite{yang2019std}                                      & Point-Voxel               & 2-stage               & $0$     & $79.8$   & -       & -       & -        & -       & -       & -        & -       & -       & -        & -        \\
		VIC-Net \cite{jiang2021vic}                                  & Point-Voxel               & 1-stage               & $0$     & $79.25$  & -       & -       & -        & -       & -       & -        & -       & -       & -        & -        \\ 
		\hline
		PointRCNN \cite{shi2019pointrcnn}                                & Point-based               & 2-stage               & $88.95$ & $78.67$  & $77.78$ & $91.83$ & $80.61$  & $78.18$ & $89.92$ & $78.67$  & $77.78$ & $93.07$ & $88.85$  & $86.73$  \\
		PointRCNN-iou \cite{shi2019pointrcnn}                            & Point-based               & 2-stage               & $89.09$ & $78.78$  & $78.26$ & $89.89$ & $80.68$  & $78.41$ & $90.19$ & $87.49$  & $85.91$ & $94.99$ & $88.82$  & $86.71$  \\
		SPSNet \cite{liang2023spsnet}                                   & Point-based               & 2-stage               & $89.19$ & $79.29$  & $78.2$  & $90.52$ & $83.03$  & $80.15$ & $90.31$ & $88.72$  & $87.31$ & $93.2$  & $91.21$  & $88.9$   \\
		3DSSD$^\mathbf{\dag}$ \cite{yang20203dssd}                                    & Point-based               & 1-stage               & $88.79$ & $78.58$  & $77.47$ & $91.32$ & $82.95$  & $80.37$ & $90.08$ & $87.87$  & $86.35$ & $93.86$ & $91.1$   & $88.78$  \\
		SASA$^\mathbf{\dag}$ \cite{chen2022sasa}                                     & Point-based               & 1-stage               & $88.88$ & $79.3$   & $78.64$ & $91.23$ & $83.09$  & $82.34$ & $90.09$ & $88.3$   & $87.19$ & $94.83$ & $91.08$  & $88.93$  \\
		IA-SSD \cite{zhang2022not}                                   & Point-based               & 1-stage               & $88.78$ & $79.12$  & $78.12$ & $89.52$ & $82.86$  & $80.05$ & $90.34$ & $88.19$  & $86.78$ & $93.17$ & $89.54$  & $88.64$  \\
		DBQ-SSD \cite{yang2022dbq}                                  & Point-based               & 1-stage               & $88.56$ & $78.74$  & $77.45$ & $89.95$ & $81.31$  & $77.14$ & $89.96$ & $87.8$   & $85.64$ & $93.49$ & $88.42$  & $87.36$  \\
		SGCCNet  & Point-based               & 1-stage               & $\uline{\textbf{89.92}}$ & $\uline{\textbf{80.8}}$   & $\textbf{78.97}$ & $\uline{\textbf{93.27}}$ & $\uline{\textbf{84.17}}$  & $\uline{\textbf{82.92}}$ & $\textbf{90.55}$ & $\uline{\textbf{88.73}}$  & $\uline{\textbf{87.94}}$ & $\uline{\textbf{96.44}}$ & $\uline{\textbf{91.65}}$  & $\uline{\textbf{89.42}}$  \\
		 \textit{-IA-SSD*} &                           &                       & $\textit{1.14}$  & $\textit{1.68}$   & $\textit{0.85}$  & $\textit{3.75}$  & $\textit{1.31}$   & $\textit{2.87}$  & $\textit{0.21}$  & $\textit{0.54}$   & $\textit{1.16}$  & $\textit{3.27}$  & $\textit{2.11}$   & $\textit{0.78}$   \\
		 \textit{-3DSSD$^\mathbf{\dag}$}  &                           &                       & $\textit{1.13}$  & $\textit{2.22}$   & $\textit{1.5}$   & $\textit{1.95}$  & $\textit{1.22}$   & $\textit{2.55}$  & $\textit{0.47}$  & $\textit{0.86}$   & $\textit{1.59}$  & $\textit{2.58}$  & $\textit{0.55}$   & $\textit{0.64}$   \\
		\bottomrule
	\end{tabular}}
\end{table*}

\subsection{Setting}
\label{sec:4.1}
\textbf{Datasets.}
We validate our method on the KITTI dataset, which best demonstrates the advantages of point-based approaches. The KITTI dataset uses a 64-beam LiDAR, resulting in relatively dense point cloud scenes where a sparse backbone can still learn ideal features. The manually annotated range in the KITTI dataset is small, leading to a significant reduction in the number of points in the scene compared to other datasets, allowing second-order complexity FPS-related algorithms to maintain fast inference speeds in this scenario. Currently, comparisons of point-based detectors are primarily focused on the KITTI dataset. The KITTI dataset is sponsored by the Karlsruhe Institute of Technology and the Toyota Technological Institute at Chicago for research in the field of autonomous driving. The widely-used dataset contains 7481 training samples with annotations in the camera field of vision and 7518 testing samples. Following the common protocol, we further divide the training samples into a training set (3,712 samples) and a validation set (3,769 samples). Additionally, the samples are divided into three difficulty levels: easy, moderate, and hard based on the occlusion level, visibility, and bounding box size. The moderate average precision is the official ranking metric for both 3D and BEV detection on the KITTI website. It is worth noting that SGCCNet, submitted to the KITTI website for comparison, is trained on 80\% of the complete training set.

\textbf{Evaluation metrics.} 
For the KITTI scene, we evaluate the performance of each class using both the 3D and BEV average precision (AP) metric. To ensure an objective comparison, we employed both the $AP$ with 40 recall points ($AP_{40}$) and the $AP$ with 11 recall points ($AP_{11}$). Consistent with the majority of state-of-the-art methods, we utilize Intersection over Union (IoU) thresholds of 0.7, 0.5, and 0.5 for \textit{Car}, \textit{Pedestrian}, and \textit{Cyclist}, respectively. 

\textbf{Implementation details.}
SGCCNet is trained for 80 epochs with a batch size of 8 on 2 Nvidia A40 GPUs. The first 70 epochs use saliency-guided point removal and new ground truth pooling for augmentation, while the last 10 epochs fine-tune using the initial ground truth pooling. The initial learning rate is set to 0.01, which is decayed by 0.1 at 35 and 45 epochs and updated with the one cycle policy. We use the Adam optimizer with $\beta_1=0.9$ and $\beta_2=0.85$ for optimization. The weight decay coefficient is set to 0.01, and the momentum coefficient is set to 0.9. Other data augmentation methods follow the default setting of Zhang et al \cite{zhang2022not}. In the WCE loss, $\epsilon$ is set to 0.001. In CCM, the initial object score threshold $score\_{thres}_1=0.01$, and the final object score threshold $score\_{thres}_2=0.45$. For potential missed detections, the missed sample incremental confidence $\Delta c$ is set to 0.2, the missed sample IoU threshold $iou\_thres_{missed}$ is set to 0.9, and the threshold of the number of missed sample neighbors $neighbor\_thres_{missed}$ is set to 10. All experiments are implemented using the OpenPCDet framework \footnote{\url{https://github.com/open-mmlab/OpenPCDet}}. The SGCCNet-elite for classification tasks is trained for 10 epochs on a single Nvidia 3090 GPU. The initial learning rate is set to 0.01, the weight decay coefficient is set to 0.0002, and the momentum coefficient is set to 0.9.

\textbf{Benchmark detector.}
The most similar work to our SGCCNet is IA-SSD, which provides the best balance between detection accuracy and inference speed in point-based pipelines to date. Its multiple metrics achieve state of the art on the KITTI dataset. The training details, model parameter configurations, and overall performance of SGCCNet are roughly similar to IA-SSD, and we will focus on comparing the performance of SGCCNet and IA-SSD.
\subsection{Comparison with state-of-the-art (SOTA) methods}
\label{sec:4.2}

\textbf{Note.} SGCCNet is trained on multiple classes simultaneously. For models trained on single classes only, they are marked as \textit{MODEL$^\mathbf{\dag}$} in the table. The IA-SSD model has sparked widespread research interest since its release, but many researchers have been unable to reproduce the performance reported in the original paper, and the authors have not responded to these concerns \footnote{\url{https://github.com/yifanzhang713/IA-SSD/issues/54}}. We believe that the IA-SSD model may rely on the training environment, especially the versions of Pytorch and CUDA. To ensure fair comparisons, we trained and tested the model in our local environment. The reproduced model is marked as \textit{MODEL$^\mathbf{*}$} in the table.

\begin{table*}
	\renewcommand\arraystretch{.6}
	\centering
	\setlength{\extrarowheight}{0pt}
	\addtolength{\extrarowheight}{\aboverulesep}
	\addtolength{\extrarowheight}{\belowrulesep}
	\setlength{\aboverulesep}{0pt}
	\setlength{\belowrulesep}{0pt}
	\caption{Quantitative comparison other point-based methods on the KITTI \textit{val} set for \textit{Cyclist} and \textit{Pedestrian} 3D detection, under the evaluation metric of 3D Average Precision ($AP$) of 11and 40 sampling recall points. Our SGCCNet results are highlighted in \textbf{Bold}.}
	\resizebox{\textwidth}{!}{
	\begin{tabular}{llll|lll|lll|lll} 
		\toprule
		\multirow{3}{*}{Method}                    & \multicolumn{6}{l|}{3D@Cyclist(IoU=0.5)}         & \multicolumn{6}{l}{3D@Pedestrian(IoU=0.5)}        \\ 
		\cline{2-13}
		& \multicolumn{3}{l|}{R11}     & \multicolumn{3}{l|}{R40}     & \multicolumn{3}{l|}{R11}     & \multicolumn{3}{l}{R40}       \\
		& Easy    & Moderate & Hard    & Easy    & Moderate & Hard    & Easy    & Moderate & Hard    & Easy    & Moderate & Hard     \\ 
		\hline
		3DSSD$^\mathbf{\dag}$                                      & $86.36$ & $71.2$   & $66.1$  & $91.4$  & $71.8$   & $67.6$  & $55.9$  & $50.62$  & $47.88$ & $55.1$  & $50.63$  & $46.2$   \\
		IA-SSD*                                     & $86.33$ & $69.13$  & $65.32$ & $89.02$ & $69.3$   & $65.29$ & $61.9$  & $57.8$   & $52.58$ & $61.06$ & $56.6$   & $51.84$  \\
		DBQ-SSD*                                    & $85.86$ & $70.15$  & $66.43$ & $90.24$ & $70.9$   & $66.12$ & $59.73$ & $54.71$  & $50.32$ & $59.26$ & $54.23$  & $48.51$  \\
		SGCCNet   & $\textbf{86.52}$ & $\textbf{72.23}$  & $\textbf{66.32}$ & $\textbf{90.84}$ & $\textbf{71.2}$   & $\textbf{66.73}$ & $\textbf{62.78}$ & $\textbf{58.1}$   & $\textbf{52.5}$  & $\textbf{62.06}$ & $\textbf{56.84}$  & $\textbf{51.73}$  \\
		\textit{-3DSSD$^\mathbf{\dag}$}   & $\textit{0.16}$  & $\textit{1.03}$   & $\textit{0.22}$ & $\textit{-0.56}$ & $\textit{-0.6}$   & $\textit{-0.87}$ & $\textit{6.88}$  & $\textit{7.48}$   & $\textit{4.62}$  & $\textit{6.96}$  & $\textit{6.21}$   & $\textit{5.53}$   \\
		\textit{-IA-SSD*}  & $\textit{0.19}$  & $\textit{3.1}$    & $\textit{1}$  & $\textit{1.82}$  & $\textit{1.9}$    & $\textit{1.44}$  & $\textit{0.88}$  & $\textit{0.3}$    & $\textit{-0.08}$ & $\textit{1}$     & $\textit{0.24}$   & $\textit{-0.11}$  \\
		\textit{-DBQ-SSD*} & $\textit{0.66}$  & $\textit{2.08}$   & $\textit{-0.11}$ & $\textit{0.6}$   & $\textit{0.3}$    & $\textit{0.61}$  & $\textit{3.05}$  & $\textit{3.39}$   & $\textit{2.18}$  & $\textit{2.8}$   & $\textit{2.61}$   & $\textit{3.22}$   \\
		\bottomrule
	\end{tabular}}
	\label{tab:3}
\end{table*}

\textbf{Performance on KITTI \textit{test} set.}
We compared the performance of SGCCNet with other SOTA models on the KITTI \textit{test} set as shown in Table. \ref{tab:1}. SGCCNet achieves an $AP_{3D}^{40}$ of 80.82\% on the \textit{moderate} level for \textit{Car} class objects, which is the best among all point-based detectors. In detail, SGCCNet outperforms IA-SSD by $0.9\%, 0.69\%, 0.54\%$ in $AP_{3D}^{40}$ on \textit{easy}, \textit{moderate}, and \textit{hard} levels respectively, and by $0.78\%, 0.55\%, 0.92\%$ in $AP_{BEV}^{40}$. Compared to the reproduced IA-SSD$^\mathbf{\dag}$, SGCCNet surpasses it by $2.1\%, 2.35\%, 2.05\%$ in $AP_{3D}^{40}$ and by $1.36\%, 1.17\%, 1.55\%$ in $AP_{BEV}^{40}$. Surprisingly, SGCCNet's performance even exceeds some structure-based models, with $mAP_{3D}^{40}$ surpassing Fast Point R-CNN by approximately $4.26\%$ and Part-A$2$ by about $1.94\%$.

\textbf{Performance on KITTI \textit{val} set.}
We further provide the results of the KITTI \textit{validation} set to better present the detection performance of our SGCCNet, as shown in Table. \ref{tab:2}. SGCCNet remains the best performing point-based detector, especially in the \textit{Easy} category, with $AP_{3D}^{40}$ exceeding IA-SSD and 3DSSD by approximately $3.75\%$ and $1.95\%$, respectively. We attribute this improvement to the confidence calibration mechanism, which allows the model to no longer rely solely on semantic scores. We also compared the performance of SGCCNet with other state-of-the-art point-based models on the \textit{Cyclist} and \textit{Pedestrian} classes, as shown in Table. \ref{tab:3}. Compared to the single-class trained 3DSSD$^\mathbf{\dag}$, SGCCNet outperforms it by $1.03\%$ and $7.48\%$ in the \textit{moderate} level of these two classes, respectively. Although its accuracy is slightly lower than 3DSSD in the \textit{Hard} class, our model is trained end-to-end for multiple classes. Compared to other multi-class trained point-based detectors such as IA-SSD and DBQ-SSD, SGCCNet outperforms them by $3.1\%$ and $2.08\%$ in the \textit{Moderate} level of \textit{Cyclist} class, and by $0.24\%$ and $2.61\%$ in the \textit{Moderate} level of \textit{Pedestrian} class. These results demonstrate that SGCCNet also excels in detecting small objects.

\begin{figure*}[t]
	\begin{center}
		\includegraphics[width=1\textwidth]{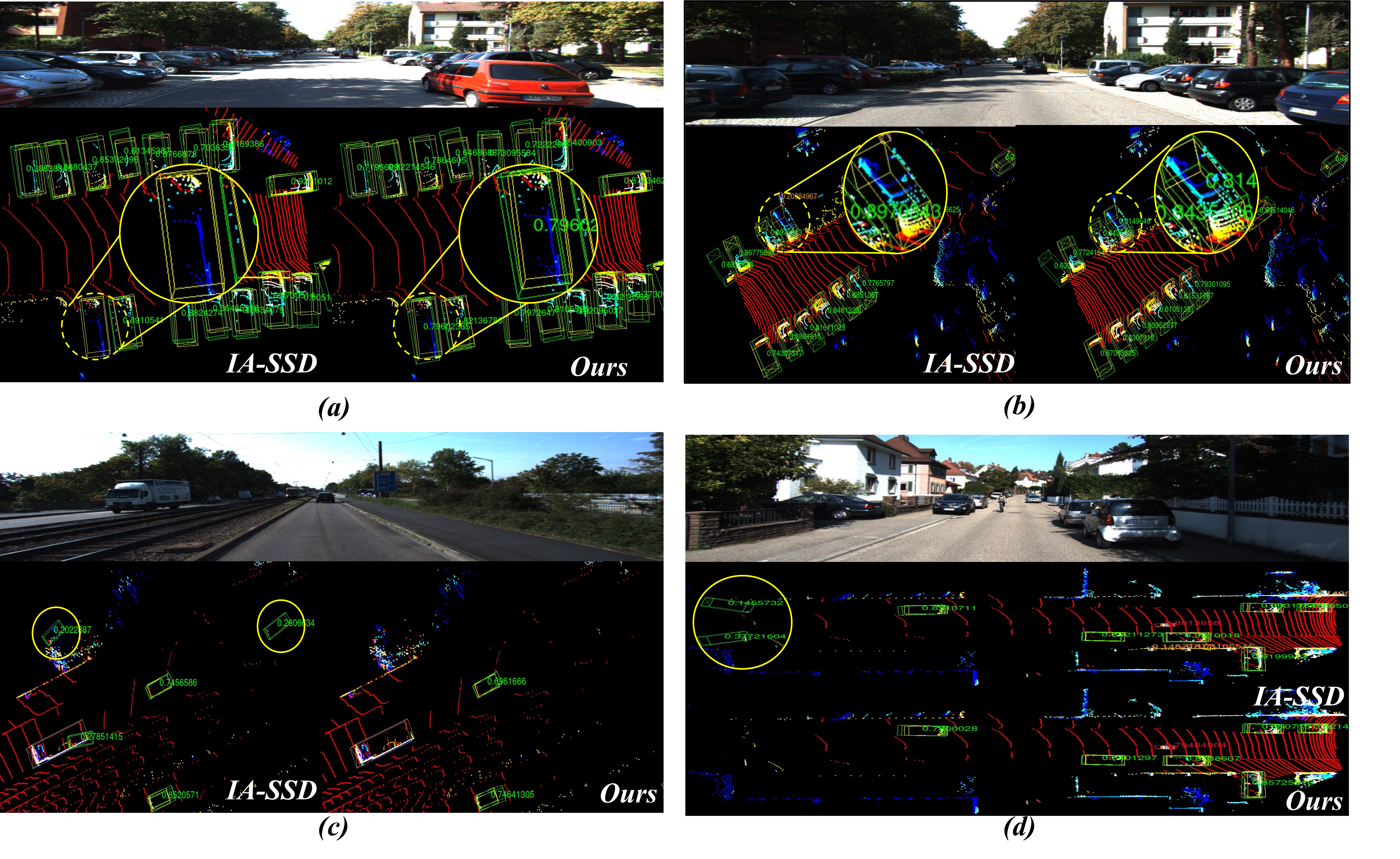}
	\end{center}
	\vspace{-0.3cm}
	\caption{Visualize the detection results of SGCCNet on the KITTI dataset. We compare the results of SGCCNet with IA-SSD, highlight the role of SGCCNet in correcting the confidence of missed detection targets and filtering false positive targets.}
	\label{fig10}
	\vspace{-0.3cm}
\end{figure*}

\textbf{Visualization.}
We quantitatively demonstrated the superiority of our SGCCNet by explaining all the above indicators. Additionally, we qualitatively illustrated the results through visualization. In Figure 8, we show four scenes from the KITTI \textit{val} set and compare the detection results of IA-SSD and SGCCNet. The yellow boxes represent ground truth (GT), while the green boxes represent the predicted results. We highlighted the areas where the predictions of the two detectors differ. In detail, in scenes (a) and (b), there is a missed detection by IA-SSD for each scene, which are high-quality targets with obvious \textit{Car} features. We believe that the reason for the missed detections is the interference of features from other objects in the region, causing the geometry and semantic information of the targets to be overlooked. However, SGCCNet can accurately detect these targets. In scenes (c) and (d), IA-SSD detected numerous false positive targets because it only considers semantic scores as confidence criteria during post-processing. In contrast, SGCCNet adjusts confidence by considering the predictions of neighboring vote points, filtering out these false positive targets. Overall, in these scenes, SGCCNet can accurately detect each target in the GT, providing intuitive evidence of the superiority of our method.

\begin{table}
	\renewcommand\arraystretch{.6}
	\centering
	\setlength{\extrarowheight}{0pt}
	\addtolength{\extrarowheight}{\aboverulesep}
	\addtolength{\extrarowheight}{\belowrulesep}
	\setlength{\aboverulesep}{0pt}
	\setlength{\belowrulesep}{0pt}
	\caption{Compare the runtime performance of models based on GPU occupancy, single-frame inference time, and data requirements. All other models are reproduced from the configuration files and weight files provided by the OpenPCD project.}
	\begin{tabular}{lllll} 
		\toprule
		Method                                   & Mem    & Time  & Input scale          & Mod.     \\ 
		\hline
		PointPliiars                             & 354MB  & $21$  & 2\textasciitilde{}9k & $77.28$  \\
		SECOND                                   & 713MB  & $33$  & 11-17k               & $78.62$  \\
		3DSSD                                    & 521MB  & $92$  & $16384$              & $78.58$  \\
		PointRCNN                                & 567MB  & $94$  & $16384$              & $78.67$  \\
		Part-A2                                  & 720MB  & $90$  & 11-17k               & $79.41$  \\
		PV-RCNN                                  & 1233MB & $126$ & 11-17k               & $79.16$  \\
		IA-SSD                                   & 102MB  & $16$  & $16384$              & $79.12$  \\
		\rowcolor[rgb]{0.855,0.89,0.953} SGCCNet & 217MB  & $23$  & $16384$              & $80.8$   \\
		\bottomrule
	\end{tabular}
\label{tab:4}
\end{table}

\subsection{Runtime Analysis}
\label{sec:4.3}

One major advantage of Point-based detectors is real-time detection. In this section, we will illustrate this point by comparing the metrics of GPU memory usage, single-frame inference time, and input data volume. Table. \ref{tab:4} shows these metrics for several representative models. It is worth noting that since their original papers did not analyze this aspect of performance, and the metrics depend on the software and hardware environment, we retested them locally using the OpenPCD project to ensure a fair comparison. The configuration files and pre-trained weights for these models all come from the OpenPCD project. The hardware used for runtime testing was a single NVIDIA RTX 3090 GPU with an Intel i7-12700KF CPU@3.6GHz, and the software used python=3.9.0 and pytorch=2.1.0+cu118. From the results, SGCCNet's memory usage is only slightly higher than IA-SSD, although the single-frame inference time is 7ms longer, it can bring a performance gain of 1.68\% $AP_{3D}^{11}$ under the same input scale, which is undoubtedly worth it. Moreover, the 23ms single-frame inference time meets the real-time requirements of current LiDAR. PointPillars can achieve a similar inference time to SGCCNet in testing, but it occupies twice the memory and has a performance decrease of 3.52\% $AP_{3D}^{11}$.

\begin{table}
	\centering
	\caption{The basic information of the created classification dataset.}
	\begin{tabular}{l|ll|l} 
		\toprule
		Split                       & \textcolor[rgb]{0.2,0.2,0.2}{Category} & Number  & \textcolor[rgb]{0.2,0.2,0.2}{Total~}  \\ 
		\hline
		\multirow{3}{*}{Train}      & Car                                    & $13382$ & \multirow{3}{*}{$16239$}              \\
		& Pedestrian                             & $2159$  &                                       \\
		& Cyclist                                & $698$   &                                       \\ 
		\hline
		\multirow{3}{*}{Validation} & Car                                    & $13874$ & \multirow{3}{*}{$17014$}              \\
		& Pedestrian                             & $2260$  &                                       \\
		& Cyclist                                & $880$   &                                       \\
		\bottomrule
	\end{tabular}
\label{tab:5}
\end{table}

\begin{figure}[t]
	\begin{center}
		\includegraphics[width=3.5in]{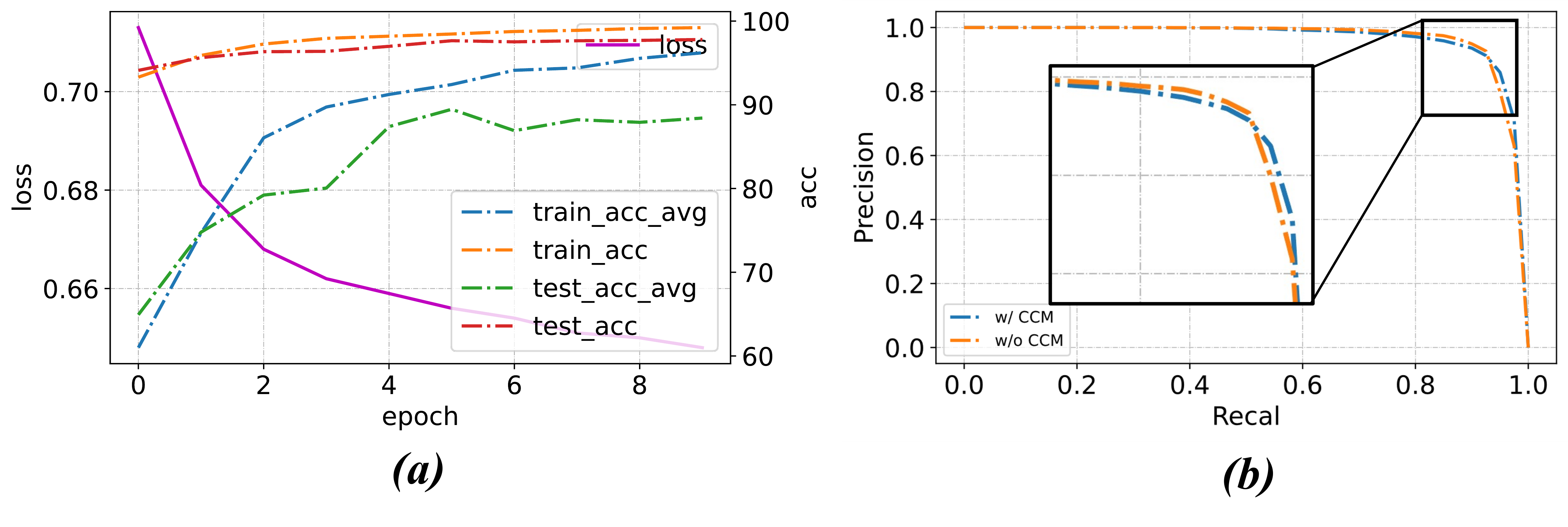}
	\end{center}
	\vspace{-0.3cm}
	\caption{(a) The classification results of SGCCNet-elite on the created dataset. (b) The PR curve of the detection task on the KITTI \textit{val} set before and after adding CCM.}
	\label{fig11}
	\vspace{-0.3cm}
\end{figure}

\begin{figure*}[t]
	\begin{center}
		\includegraphics[width=1\textwidth]{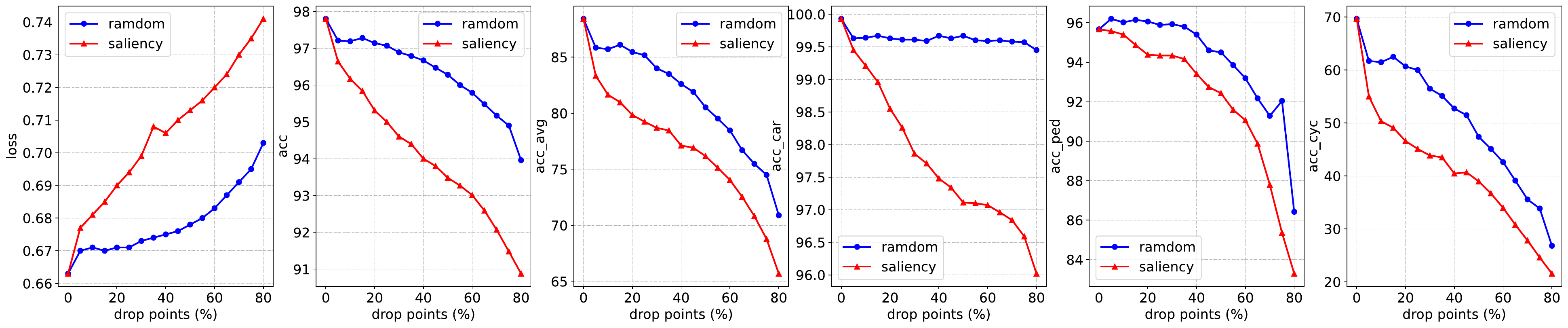}
	\end{center}
	\vspace{-0.3cm}
	\caption{Validate the saliency results on the classification task. The disruption to the model's classification performance from discarding points based on saliency scores is much greater than randomly discarding points, with higher saliency scores resulting in greater disruption.}
	\label{fig12}
	\vspace{-0.3cm}
\end{figure*}

\begin{figure}[t]
	\begin{center}
		\includegraphics[width=3.5in]{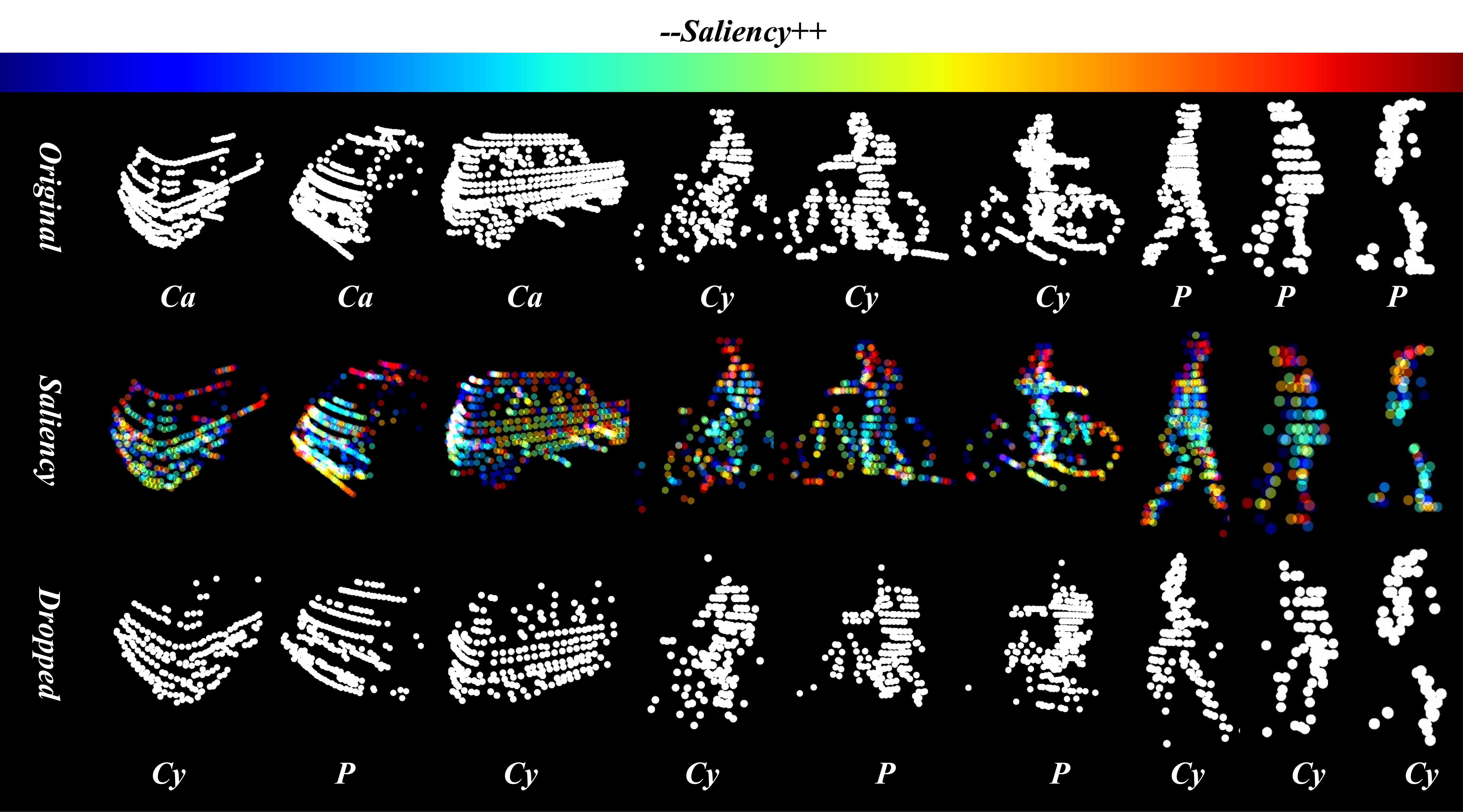}
	\end{center}
	\vspace{-0.3cm}
	\caption{Visualize the saliency heatmap. Even after discarding some points based on saliency scores, the point cloud still retains the characteristics of their respective categories. However, the model misclassifies them as other categories. This indicates that the model relies on saliency features, which is detrimental to the model's robustness. Our goal is to let the model learn from the discarded points and improve the diversity of training data at the feature level. In the figure, \textit{Ca}, \textit{Cy}, and \textit{P} respectively represent \textit{Car}, \textit{Cyclist}, and \textit{Pedestrian}.}
	\label{fig13}
	\vspace{-0.3cm}
\end{figure}

\begin{table}
	\renewcommand\arraystretch{.6}
	\centering
	\caption{The results of the ablation experiment. It can be seen that the model containing all components achieves the best performance.}
	\begin{tabular}{ccccc|ccc} 
		\toprule
		Model & SGDA & GNM & SCB & CCM & E       & M       & H        \\
		\hline
		A     &     &   &    &    & $89.52$ & $82.86$ & $80.05$  \\
		B     & \CheckmarkBold    &   &    &    & $92.11$ & $83.5$  & $82.1$   \\
		C     & \CheckmarkBold    & \CheckmarkBold  &    &    & $92.53$ & $83.65$ & $82.41$  \\
		D     & \CheckmarkBold    & \CheckmarkBold  & \CheckmarkBold   &    & $92.9$  & $83.8$  & $82.4$   \\
		E     & \CheckmarkBold    & \CheckmarkBold  & \CheckmarkBold   & \CheckmarkBold   & $\textbf{93.27}$ & $\textbf{84.17}$ & $\textbf{82.92}$  \\
		\bottomrule
	\end{tabular}
	\label{tab:6}
\end{table}

\begin{figure*}[t]
	\begin{center}
		\includegraphics[width=1\textwidth]{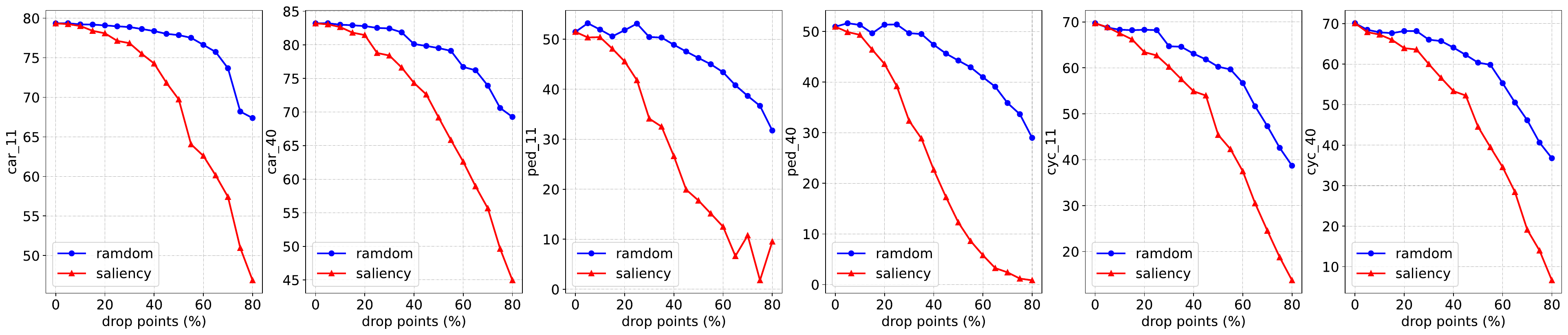}
	\end{center}
	\vspace{-0.3cm}
	\caption{Validate the saliency results on the KITTI dataset detection task. The impact of discarding points based on their saliency score on the model's detection performance is much greater than randomly discarding points. This demonstrates that saliency obtained from classification tasks is equally applicable to detection tasks, validating our previous hypothesis.}
	\label{fig14}
	\vspace{-0.3cm}
\end{figure*}

\subsection{Ablation Study}
\label{sec:44}

In this section, we will analyze the roles of various model components in SGCCNet and the selection of hyperparameters. The models involved are all trained on the KITTI \textit{train} set and tested on the \textit{val} set.

\textbf{Effect of Saliency-Guide Data Augmentation (SGDA).}
The purpose of SGDA is to increase the diversity of training data at the feature level, reduce the model's reliance on saliency features, and improve its robustness to low-quality targets. As described in Section \ref{sec:3.3}, SGDA constructs an SGCCNet-elite model to perform classification tasks and obtain saliency scores for each target. We first need to demonstrate that the saliency scores obtained from classification are also applicable to detection tasks. The data set partitioning of the training set we created is the same as KITTI, with specific sample details shown in Table. \ref{tab:5}.

Fig. \ref{fig11} shows the training and testing results of SGCCNet-elite on the classification dataset. It achieves good classification results, with an accuracy of $97.8\%$ and an average accuracy of $88.42\%$. The classification accuracies for \textit{Car}, \textit{Pedestrian}, and \textit{Cyclist} are $99.93\%$, $95.67\%$, and $69.67\%$, respectively.
To apply the method for obtaining point-wise saliency score as described in Section \ref{sec:3.3} to SGCCNet-elite, we conducted a dropout experiment as shown in Fig. \ref{fig12}. We gradually removed salient points (SD) from $5\%$ to $80\%$ of the samples according to Algorithm. \ref{Alg:alg1}, and observed the changes in classification metrics, while also comparing with random dropout (RD). As shown in the Fig. \ref{fig12}, as the number of dropped points increases, both SD and RD methods decrease the classification performance of the model. However, SD is more destructive to the model, with a faster decline in performance. Even when only 5\% of points are dropped, the classification accuracy for the \textit{Cyclist} class decreased by 14.67\%. This comparison demonstrates that the significance scores we obtained are suitable for classification tasks.
In Fig. \ref{fig13}, we list some samples for saliency analysis, showing that for the same type of target, the distribution of saliency regions is similar. For example, \textit{Car} targets are concentrated on the roof position, while \textit{Pedestrian} and \textit{Cyclist} targets are concentrated on the human body position. The model's dependence on saliency features reduces its robustness. We discard some saliency points, although still clearly retaining the basic features of these targets, the model misclassifies them as other categories. Our SGDA aims to alleviate this issue.

Similarly, we applied this dropout experiment on the targets in each scene of the KITTI \textit{val} set for the detection task, and the experimental results are shown in Fig. \ref{fig14}. It can be seen that SD significantly decreases the detection performance of each class of targets, and the decrease rate is much higher than RD. We believe that such experimental results are sufficient to demonstrate that the saliency obtained from the classification task is also applicable to the detection task. By using this saliency-guided data augmentation method to discard salient region features, it differs from previous random augmentation methods and can truly enhance the diversity of training data at the feature level with a specific purpose. Table. \ref{tab:6} quantitatively illustrates the performance gain brought by SGDA, which can lead to a $1.76\% mAP_{3D}^{11}$ improvement. The above results intuitively demonstrate the effectiveness of SGDA.

\begin{table}
	\renewcommand\arraystretch{.6}
	\centering
	\caption{We test GNM and SCB at different stages to determine the best combination.}
	\begin{tabular}{cc|ccc|ccc} 
		\toprule
		&                           & Stage1 & Stage2 & Stage3 & E                        & M                        & H                         \\
		\hline
		\multirow{2}{*}{A} & Norm                      &        &        &        & \multirow{2}{*}{$92.11$} & \multirow{2}{*}{$83.5$}  & \multirow{2}{*}{$82.1$}   \\
		& Skip                      &        &        &        &                          &                          &                           \\ 
		\hline
		\multirow{2}{*}{B} & Norm                      &        & \CheckmarkBold     & \CheckmarkBold    & \multirow{2}{*}{$92.53$} & \multirow{2}{*}{$83.62$} & \multirow{2}{*}{$82.31$}  \\
		& Skip                      &        & \CheckmarkBold      & \CheckmarkBold      &                          &                          &                           \\ 
		\hline
		\multirow{2}{*}{C} & Norm                      & \CheckmarkBold      & \CheckmarkBold      & \CheckmarkBold      & \multirow{2}{*}{$\textbf{92.96}$} & \multirow{2}{*}{$83.47$} & \multirow{2}{*}{$82.24$}  \\
		& Skip                      & \CheckmarkBold      & \CheckmarkBold      & \CheckmarkBold      &                          &                          &                           \\ 
		\hline
		\multirow{2}{*}{D} & Norm                      & \CheckmarkBold      & \CheckmarkBold      & \CheckmarkBold      & \multirow{2}{*}{$92.9$}  & \multirow{2}{*}{$\textbf{83.8}$}  & \multirow{2}{*}{$\textbf{82.4}$}   \\
		& Skip                      &        & \CheckmarkBold      & \CheckmarkBold      &                          &                          &                           \\ 
		\hline
		\multirow{2}{*}{E} & Norm                      &        & \CheckmarkBold      & \CheckmarkBold      & \multirow{2}{*}{$92.71$} & \multirow{2}{*}{$83.54$} & \multirow{2}{*}{$82.29$}  \\
		& \multicolumn{1}{l|}{Skip} & \CheckmarkBold      & \CheckmarkBold      & \CheckmarkBold      &                          &                          &                           \\
		\bottomrule
	\end{tabular}
	\label{tab:7}
\end{table}

\textbf{Effects of Geometric Normalization Module (GNM) and Skip Connection Block (SCB).} 
The original intention of GNM and SCB designs is to address the issues of Internal Covariate Shift and feature forgetting that may arise in models, respectively. The structures of both are often used in advanced classification models. In Table. \ref{tab:7}, we analyze the performance changes of models when GNM and SCB are placed in different positions. Adding GNM and SCB in the second and third stages both bring significant benefits to the model, resulting in a $0.42\% AP_{3D}^{11}$ improvement for the \textit{Car} class at the \textit{Easy} level. However, when GNM and SCB are added to all three stages, the performance of the model may even decrease. After conducting multiple experiments to rule out the influence of random factors on the model, we found that it may be the GNM or SCB in the first stage that damaged the feature distribution. Through experiments (models C, D in Table. \ref{tab:7}), we ultimately decided to use only GNM in the first stage and abandon SCB. The early point-wise features did not learn the essential features of the targets, and directly concatenating them with the features of the later stages may affect the predictions in the later stages of the model. As shown in Table. \ref{tab:6}, the combination of GNM and SCB can bring a $0.46\% mAP_{3D}^{11}$ improvement to the model.

\begin{table*}
	\centering
	\caption{Comparison of model performance before and after considering missed targets.}
	\begin{tabular}{l|llllll} 
		\toprule
		\multirow{3}{*}{Method} & \multicolumn{6}{l}{Car(IoU=0.7)}                             \\ 
		\cline{2-7}
		& \multicolumn{3}{l}{R11}      & \multicolumn{3}{l}{R40}       \\
		& Easy    & Moderate & Hard    & Easy    & Moderate & Hard     \\ 
		\hline
		SGCCNet w/o missed      & $89.62$ & $80.65$  & $78.78$ & $92.89$ & $83.93$  & $82.94$  \\
		SGCCNet w/ missed       & $89.92$ & $80.8$   & $78.97$ & $93.27$ & $84.17$  & $82.92$  \\
		\bottomrule
	\end{tabular}
	\label{tab:8}
\end{table*}

\begin{figure}[t]
	\begin{center}
		\includegraphics[width=2.5in]{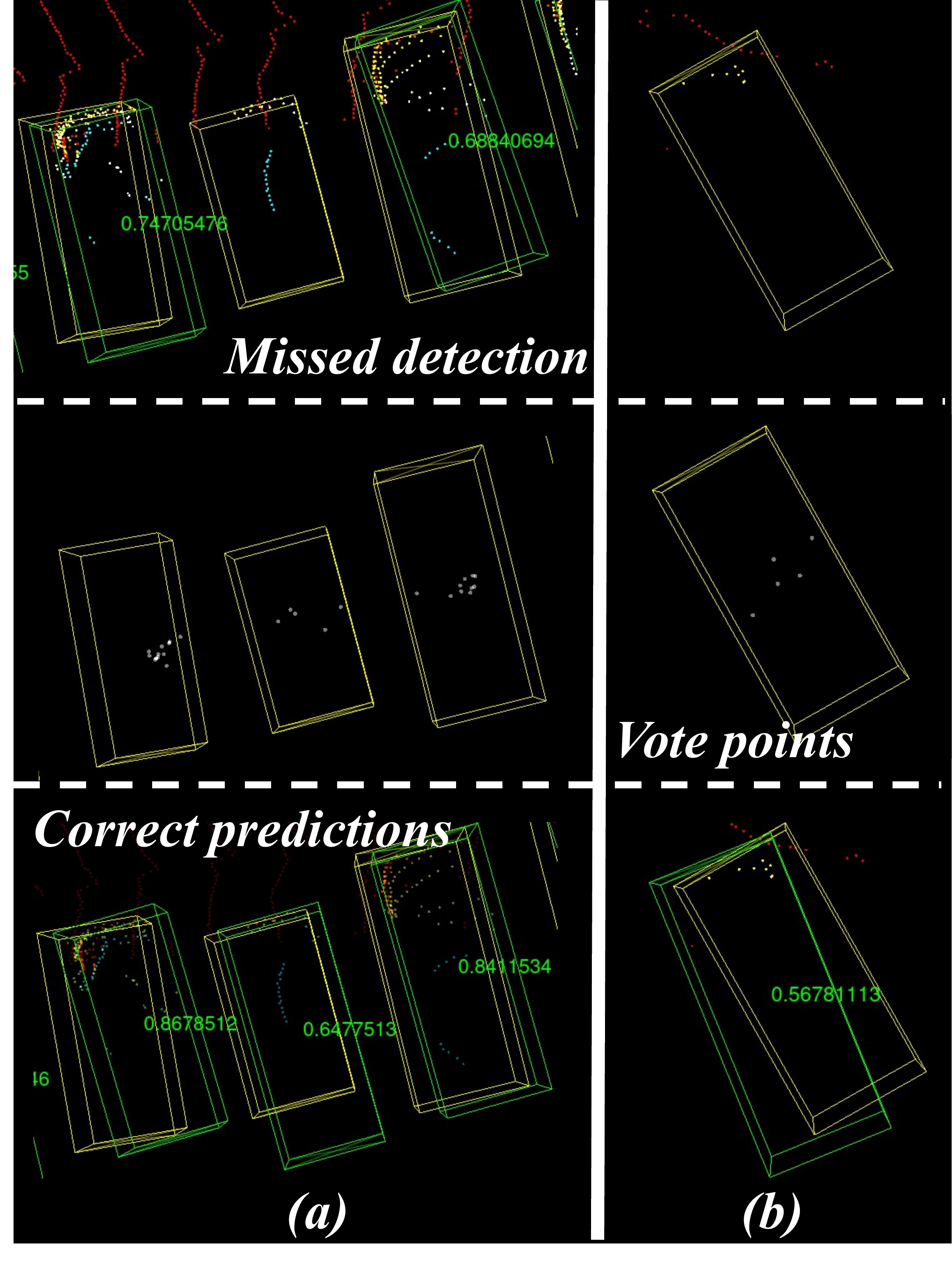}
	\end{center}
	\vspace{-0.3cm}
	\caption{Our CCM takes into account the possibility of missing targets in the model. For accurate localization predictions, we give it a certain confidence gain to ensure that it is detected by the model.}
	\label{fig15}
	\vspace{-0.3cm}
\end{figure}

\textbf{Confidence Correction Mechanism (CCM).} 
The current single-stage point-based detector tends to cause misalignment between localization accuracy and classification confidence (MLC) issue by simply using semantic scores as the sole criterion for filtering predicted boxes during post-processing. SGCCNet proposes a Confidence Calibration Module (CCM) for point-based multi-class detectors to alleviate this problem. As shown in Algorithm \ref{Alg:alg2}, the purpose of this CCM is to correct the confidence of the current vote point by using the prediction information of neighboring vote points. Table \ref{tab:6} quantitatively demonstrates the effectiveness of CCM, which can improve the model's $mAP_{3D}^{11}$ by $0.42\%$.

More intuitively, in Fig. \ref{fig11} (b), we demonstrate the Precision-Recall (PR) curve of the model before and after adopting CCM. Although CCM itself does not possess learning capabilities, it can filter out false positive boxes and boxes with high semantic scores but inaccurate localization based on the predicted geometric information. As shown in the figure, the model can maintain a higher accuracy for a longer duration at higher recall, indicating a lower number of false positive targets compared to the scenario without CCM. We also illustrate the effectiveness of our consideration for missed targets in Table. \ref{tab:8}. It can be observed that after adding a certain confidence gain to targets with accurate localization but low semantic scores, the model exhibits significant improvements in various performance metrics. Fig. \ref{fig15} visually demonstrates two such targets that possess distinct geometric information similar to the target, and they are ultimately detected by CCM.
\begin{figure}[t]
	\begin{center}
		\includegraphics[width=.5\textwidth]{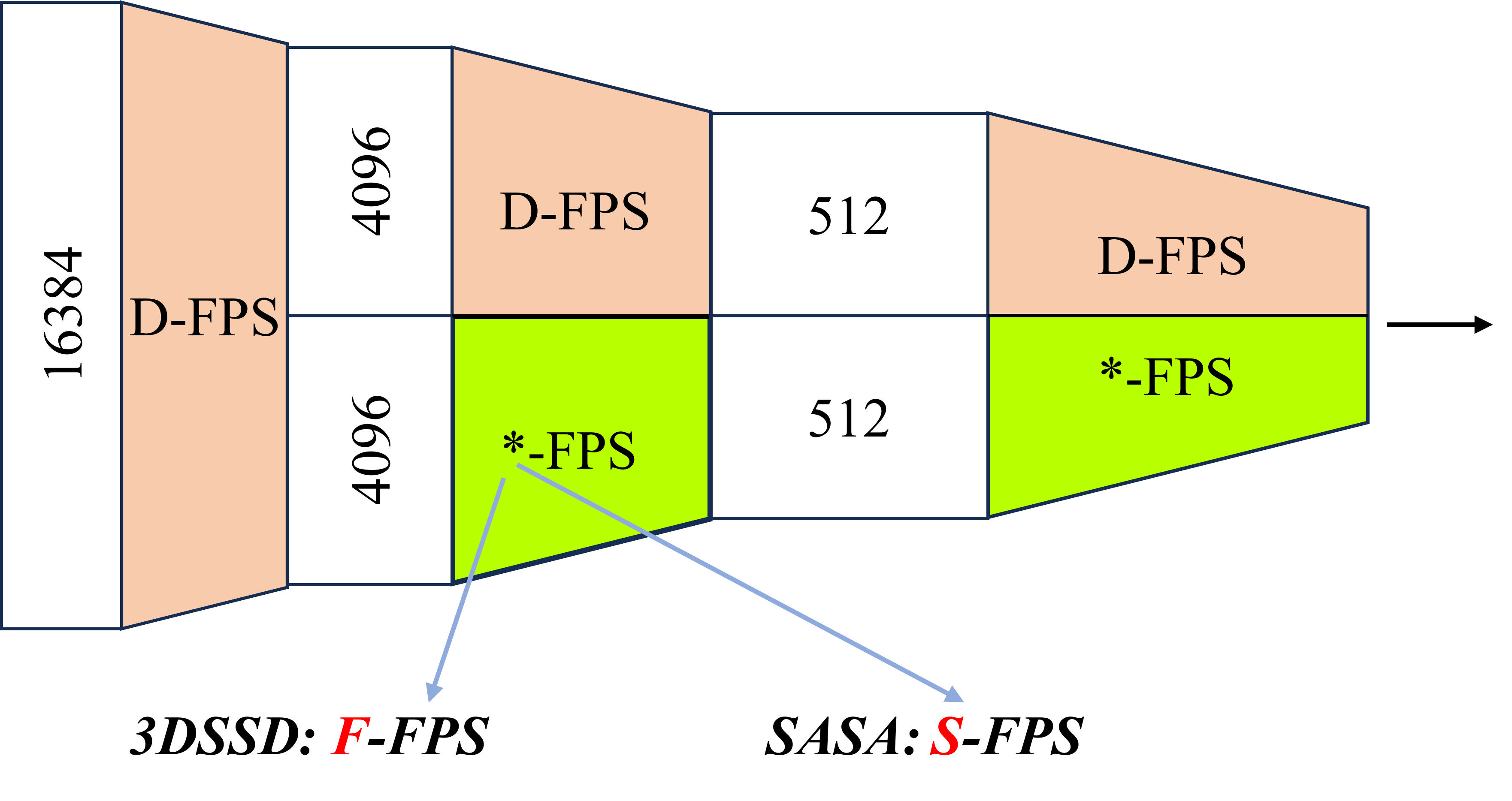}
	\end{center}
	\vspace{-0.3cm}
	\caption{The basic structure of 3DSSD and SASA. The only difference lies in the downsampling method, where the former utilizes Farthest Point Sampling based on feature distance (F-FPS), while the latter employs Semantic Weighted Farthest Point Sampling (S-FPS).}
	\label{fig16}
	\vspace{-0.3cm}
\end{figure}

\begin{table}
	\renewcommand\arraystretch{.6}
	\centering
	\setlength{\extrarowheight}{0pt}
	\addtolength{\extrarowheight}{\aboverulesep}
	\addtolength{\extrarowheight}{\belowrulesep}
	\setlength{\aboverulesep}{0pt}
	\setlength{\belowrulesep}{0pt}
	\caption{The versatility test of SGCCNet. It can be seen that SGDA and CCM can provide significant performance gains for 3DSSD and SASA.}
	\resizebox{3.5in}{!}{
	\begin{tabular}{lllllll} 
		\toprule
		\multirow{3}{*}{Method}                & \multicolumn{6}{l}{Car(IoU=0.7)}                             \\ 
		\cline{2-7}
		& \multicolumn{3}{l}{R11}      & \multicolumn{3}{l}{R40}       \\
		& Easy    & Moderate & Hard    & Easy    & Moderate & Hard     \\ 
		\hline
		3DSSD                                  & $88.79$ & $78.58$  & $77.47$ & $91.32$ & $82.95$  & $80.37$  \\
		3DSSD+SGDA                              & $89.13$ & $79.67$  & $78.33$ & $91.97$ & $83.29$  & $82.18$  \\
		\textit{Gain} & $\textit{0.34}$  & $\textit{1.09}$   & $\textit{0.86}$  & $\textit{0.65}$  & $\textit{0.34}$   & $\textit{1.81}$   \\
		3DSSD+SGDA+CCM                          & $89.65$ & $80.14$  & $78.73$ & $92.16$ & $83.65$  & $82.29$  \\
		\textit{Gain} & $\textit{0.86}$  & $\textit{1.56}$   & $\textit{1.26}$  & $\textit{0.84}$  & $\textit{0.7}$    & $\textit{1.92}$   \\ 
		\hline\hline
		SASA                                   & $88.88$ & $79.3$   & $78.64$ & $91.23$ & $83.09$  & $82.34$  \\
		SASA+SGDA                               & $89.45$ & $79.47$  & $78.77$ & $91.64$ & $83.29$  & $82.36$  \\
		\textit{Gain} & $\textit{0.57}$  & $\textit{0.17}$   & $\textit{0.13}$  & $\textit{0.41}$  & $\textit{0.2}$    & $\textit{0.02}$   \\
		SASA+SGDA+CCM                           & $89.74$ & $79.85$  & $79.01$ & $92.03$ & $83.35$  & $82.52$  \\
		Gain & $\textit{0.86}$  & $\textit{0.55}$   & $\textit{0.37}$  & $\textit{0.8}$   & $\textit{0.26}$   & $\textit{0.18}$   \\
		\bottomrule
	\end{tabular}}
	\label{tab:9}
\end{table}

\subsection{Generalizability Analysis}
\label{sec:45}
The core components of SGCCNet are plug-and-play, and we also tested their versatility on other point-based detectors. We selected the 3DSSD and SASA detectors for analysis, as they have similar basic structures as shown in Fig. \ref{fig16}, with the only difference being the downsampling method. To maintain the basic structure of the models, we only analyzed the effects of SGDA and CCM. The experimental results, as shown in Table. \ref{tab:9}, demonstrate significant improvements in the performance of both models after adding SGDA and CCM. 3DSSD shows an improvement of $0.34\%$, $1.09\%$, and $0.86\%$ in $AP^{11}{3D}$ on \textit{Easy}, \textit{Moderate}, and \textit{Hard} levels, respectively. SASA shows improvements of $0.86\%$, $0.55\%$, and $0.37\%$ respectively. These results directly indicate that SGCCNet is easy to be plugged into popular architectures.
\subsection{Discussion}
\label{sec:46}
In this section, we have detailed the superiority and effectiveness of SGCCNet from the perspectives of detection metrics, visualization, efficiency analysis, and ablation experiments. Specifically, SGCCNet is currently the best-performing multi-class single-stage end-to-end point-based detector on the KITTI dataset, achieving an $AP_{3D}^{11}$ of $80.82\%$ on the KITTI \textit{test} set, surpassing IA-SSD by approximately $2.35\%$, and even outperforming detectors that adopt structure-backbone. Additionally, SGCCNet has an inference speed of 23ms, slightly lower than IA-SSD but fully compliant with the real-time requirements of current LiDAR systems. Furthermore, the core components of SGCCNet are plug-and-play, providing significant performance gains for other similar detectors such as 3DSSD and SASA. Overall, the design of SGCCNet is based on addressing the issues of ILQ and MLC that exist in current point-based detectors. From the results, it is evident that SGCCNet effectively mitigates the impact of these issues on the model, improving the safety baseline and application prospects of point-based detectors.
\section{Conclusion}
\label{sec:5}
In this paper, we propose a new single-stage point-based 3D object detector called SGCCNet. It incorporates a saliency-guided data augmentation method that enhances the diversity of training data at the feature level, reducing the model's reliance on salient features and improving its robustness to low-quality objects. It also includes a geometric normalization module and skip connection block to address the challenges of Internal Covariate Shift and feature forgetting. To tackle the misalignment between localization accuracy and classification confidence in single-stage detectors, SGCCNet introduces a confidence calibration mechanism suitable for multi-class point-based detectors. Experimental results demonstrate that SGCCNet is currently the best-performing point-based detector, with its model components playing a substantial role in enhancing model performance and exhibiting good transferability.

\textbf{Limitations and outlook.}
Like other point-based detectors, SGCCNet performs well only in LiDAR scenes with multi-beams, such as the KITTI dataset, but its overall performance is far inferior to structure-based detectors on datasets like NuScenes \cite{nuscenes} and Waymo \cite{Sun_2020_CVPR}. Firstly, this is because these scenes are more complex, making it difficult to form a suitable fixed configuration, and the sparser distribution of objects hampers feature learning for point-based detectors. Secondly, the quadratic complexity of the FPS algorithm requires a significant amount of time in such multi-point scenes. In future work, we will develop more efficient point-based backbones to address these two issues.



\ifCLASSOPTIONcaptionsoff
  \newpage
\fi

\bibliographystyle{IEEEtran}
\bibliography{Collection}

\vfill


\end{document}